\documentclass{article}

    \usepackage[preprint]{neurips_2025}

\usepackage[utf8]{inputenc} % allow utf-8 input
\usepackage[T1]{fontenc}    % use 8-bit T1 fonts
\usepackage{hyperref}       % hyperlinks
\usepackage{url}            % simple URL typesetting
\usepackage{booktabs}       % professional-quality tables
\usepackage{amsfonts}       % blackboard math symbols
\usepackage{nicefrac}       % compact symbols for 1/2, etc.
\usepackage{microtype}      % microtypography
\usepackage{xcolor}         % colors
\usepackage{multirow}
\usepackage{natbib}
\usepackage{wrapfig}
\usepackage{adjustbox}

\title{A Set-to-Set Distance Measure in Hyperbolic Space}

\usepackage[utf8]{inputenc} % allow utf-8 input
\usepackage[T1]{fontenc}    % use 8-bit T1 fonts
\usepackage{hyperref}       % hyperlinks
\usepackage{url}            % simple URL typesetting
\usepackage{booktabs}       % professional-quality tables
\usepackage{amsfonts}       % blackboard math symbols
\usepackage{nicefrac}       % compact symbols for 1/2, etc.
\usepackage{microtype}      % microtypography
\usepackage{xcolor}         % colors

\usepackage{graphicx}   
\usepackage{amsmath}
\usepackage{amssymb}
\usepackage{booktabs}
\usepackage{acronym}
\usepackage{cleveref}

\usepackage{algorithm}
\usepackage{algorithmic}
\usepackage{multirow}
\usepackage{lipsum}

\usepackage{caption}
\usepackage{subcaption}  % Modern replacement for subfigure

\usepackage{tcolorbox}

\newtheorem{theorem}{Theorem}[section]
\newtheorem{proposition}[theorem]{Proposition}
\newtheorem{proof}[theorem]{Proof}
\newtheorem{lemma}[theorem]{Lemma}
\newtheorem{corollary}[theorem]{Corollary}
\newtheorem{definition}[theorem]{Definition}

\newtheorem{remark}[theorem]{Remark}

\usepackage{wrapfig}

\author{
Pengxiang Li$^{1}$\thanks{Equal contribution.}~~~Wei Wu$^{1*}$
\textbf{Zhi Gao}$^{1}$, \textbf{Xiaomeng Fan}$^{1}$ \textbf{Peilin Yu}$^{1}$  \\ \textbf{Yuwei Wu}$^{1,2}$ \textbf{Zhipeng Lu}$^{2}$
\textbf{Mehrtash Harandi}$^{3}$
\textbf{Yunde Jia}$^{2,1}$ \\
\small $^1$Beijing Key Laboratory of Intelligent Information Technology, \\ \small School of Computer Science \& Technology, Beijing Institute of Technology \\
\small $^2$Guangdong Laboratory of Machine Perception and Intelligent Computing, Shenzhen MSU-BIT University \\
\small $^3$ Department of Electrical and Computer System Engineering, Monash University \\
  }
\begin{document}

\maketitle

\begin{abstract}
We propose a hyperbolic set-to-set distance measure for computing dissimilarity between sets in hyperbolic space. While point-to-point distances in hyperbolic space effectively capture hierarchical relationships between data points, many real-world applications require comparing sets of hyperbolic data points,
where the local structure and the global structure of the sets carry crucial semantic information. 
The proposed the \underline{h}yperbolic \underline{s}et-\underline{to}-\underline{s}et \underline{d}istance measure (HS2SD) integrates both global and local structural information: global structure through geodesic distances between Einstein midpoints of hyperbolic sets, and local structure through topological characteristics of the two sets.
To efficiently compute topological differences, we prove that using a finite Thue-Morse sequence of degree and adjacency matrices can serve as a robust approximation to capture the topological structure of a set.
In this case, by considering the topological differences, HS2SD provides a more nuanced understanding of the relationships between two hyperbolic sets.
Empirical evaluation on entity matching, standard image classification and few-shot image classification demonstrates that our distance measure outperforms existing methods by effectively modeling the hierarchical and complex relationships inherent in hyperbolic sets. 
\end{abstract}

\vspace{-10pt}
\section{Introduction}
\label{intro}

% Hyperbolic spaces can be viewed as the continuous counterpart of discrete trees~\citep{pmlr-v80-sala18a,balazevic2019multi}, providing significant advantages for modeling data with implicit or explicit treelike structures\citep{krioukov2010hyperbolic, sala2018representation}. 
% 

Learning algorithms in hyperbolic spaces with constant negative curvature have achieved impressive performance across multiple domains, including few-shot learning~\citep{gao2021curvature,ijcai2022p517},
clustering~\citep{Lin_2023_ICCV}, retrieval~\citep{Ermolov_2022_CVPR}, segmentation~\citep{Atigh_2022_CVPR}, multi-modal learning~\citep{,Long_2023_ICCV}, and 3D vision~\citep{hsu2021capturing}.
% , montanaro2022rethinking,lin2023hyperbolic}.
The reason behind their success is mainly due to the exponentially growing space volume in hyperbolic spaces with the radius~\citep{pmlr-v80-sala18a,balazevic2019multi}, providing minimal distortion for hierarchical structures of practical data.%~\citep{sarkar2011low}.

% In the current literature, most existing algorithms use point-to-point hyperbolic distance measures for downstream tasks ~\citep{khrulkov2020hyperbolic,long2020searching,atigh2022hyperbolic}. They extract a global representation of each data sample, and compute the geodesic distance between samples for classification~\citep{ye2020fewshot}, segmentation~\citep{chen2023hyperbolic}, and retrieval~\citep{yan2023hyp}.
% However, many practical tasks require formulating a set-to-set hyperbolic distance. 
% For example, in the entity matching task~\citep{hui2022localized}, each entity has multiple attributes, and the goal is to determine whether two records refer to the same real-world entity, as shown in \cref{fig:teaser} (a).
% It is desirable to model an entity as each tuple in the table represents an entity with order-agnostic attributes, \zhi{???}
% and use a hyperbolic set-to-set distance to explore the inherent relationships among attributes for entity matching.
% Additionally, in the image few-shot learning task~\citep{zhang2022hyperbolic,hamzaoui2024hyperbolic}, 
% constructing the feature pyramid using features from different layers of neural networks increases the discriminative power, as shown in \cref{fig:teaser} (b). It is natural to model one image as a set containing features from different layers and use a set-to-set distance to classify images.
% Unfortunately, the set-to-set distance measure in hyperbolic spaces has yet to be explored.

\begin{figure}
    \centering
    \includegraphics[width=0.9\linewidth]{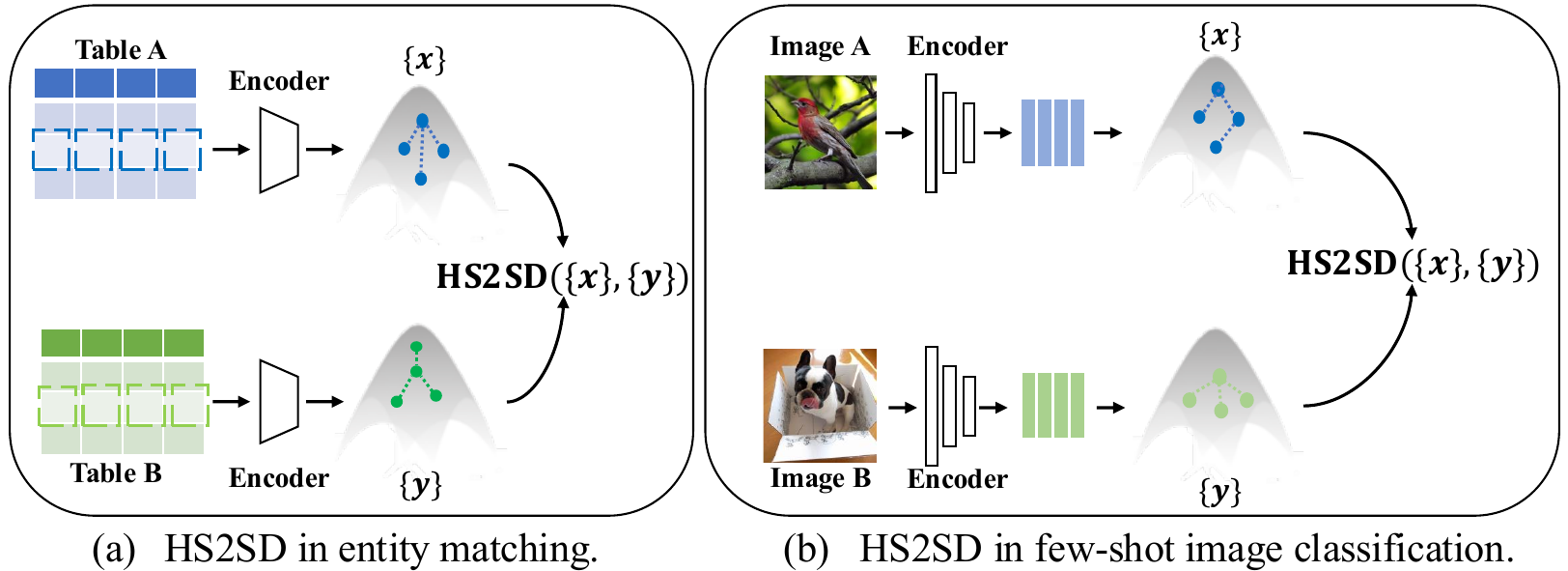}
    % \vspace{-17px}
    \caption{Examples of HS2SD on the entity matching and few-shot learning tasks.}
    \label{fig:teaser}
    \vspace{-15pt}
\end{figure}

In the current literature, hyperbolic geometry is usually exploited via point-to-point distance measures: each sample is embedded as a single data point, and downstream tasks such as classification~\citep{ye2020fewshot}, segmentation~\citep{chen2023hyperbolic}, and retrieval~\citep{yan2023hyp} are performed by computing geodesic distances between the data points~\citep{khrulkov2020hyperbolic,long2020searching,atigh2022hyperbolic}.  
However, many real-world problems naturally involve comparisons between sets of hyperbolic embeddings rather than individual points.  
For example, in entity matching~\citep{hui2022localized} (Fig.~\ref{fig:teaser} (a)), each record comprises an unordered set of attribute embeddings, and a hyperbolic set-to-set distance can capture the relational patterns among attributes to decide whether two records refer to the same entity.  
Similarly, in few-shot image classification~\citep{zhang2022hyperbolic,hamzaoui2024hyperbolic} (Fig.~\ref{fig:teaser} (b)), treating the multi-layer feature pyramid of a single image as a set enables richer representations, requiring a robust set-to-set distances.  
Unfortunately, the set-to-set distance measure in hyperbolic spaces has yet to be explored.

In this paper, we propose the \underline{h}yperbolic \underline{s}et-\underline{to}-\underline{s}et \underline{d}istance (HS2SD) that leverages both the global and local structures of hyperbolic sets.
% We argue that leveraging both the global and local structures is essential for hyperbolic set-to-set distance measures. 
Given two sets of hyperbolic representations, the global structures are responsible for measuring the relative positioning between them in the entire hyperbolic space, considering their hierarchical relationships. 
On the other hand, the local structures capture the topological differences between the two sets, allowing the distance measure to discern subtle variations in the sets, which might be overlooked by a purely global perspective.
By combining both global and local structures, the hyperbolic set-to-set distance enables a more nuanced and accurate measurement of dissimilarity, capable of capturing both the overall distribution of the sets and their detailed, underlying patterns. This dual consideration enhances the ability to apply set-to-set distance measures in complex tasks where both global relationships and local topological features are important.

In doing so, the HS2SD is designed to integrate two key components: geodesic distance and topological distance for the global and local structures, respectively. 
The geodesic distance aggregates hyperbolic representations of a set into an Einstein midpoint~\citep{gulcehre2018hyperbolic} and computes the hyperbolic $l_1$ distance between two Einstein midpoints for a global difference, thereby quantifying their overall dissimilarity.
The topological distance constructs a topology graph of a set and assesses the structural similarities between the topology graphs between two sets for local differences.
In this case, 
unlike existing point-to-point hyperbolic distance measures that focus on relationships between individual points, HS2SD integrates both global and local perspectives to provide a more nuanced understanding of the relationships between data sets.
% The core idea of the proposed method lies in effectively capturing the local geometric structure while preserving global similarities. To this end, we propose a novel approach that incorporates a geodesic distance measure and a geometric distance measure. The geodesic distance measure is designed to capture global similarities by aggregating a set into a representative global point and computing the distance between the global points of two sets, thereby quantifying their overall dissimilarity. To measure the local geometric structure, we introduce a geometric distance measure to quantify the structural similarity between sets. Theoretically, through the lens of isomorphism, we demonstrate that any polynomial combination of a tree's degree matrix and adjacency matrix serves as a valid descriptor for evaluating the structural similarity of trees.

% Video person Reid

To achieve this, we must solve one critical problem: how to represent the topological structures of the sets and compute their differences.
In doing so, we construct a topology graph based on the distances between points and compute its adjacency and degree matrices. 
Theoretically, through the lens of isomorphism ~\citep{McKay}, we demonstrate that the infinite polynomial combination of degree matrix and adjacency matrix serves as a valid descriptor for evaluating the structural similarity of trees.
To make a feasible and efficient computation of the infinite polynomial combinations, we 
% thue1912gegenseitige,
 model the combinations of degree and adjacency matrices leveraging Thue-Morse sequences~\citep{morse1921recurrent}, which exhibit uniform distributions to well approximate the infinite polynomial combination and are computed recursively to reduce the computation burdens~\citep{kuipers2012uniform,allouche2003automatic}.
By computing the Frobenius norm of the difference between two Thue-Morse sequences, we derive the topological distance.
We conduct experiments on the entity matching and few-shot image classification tasks, and our method achieves better performance than compared methods that use a point-to-point distance measure, showing the effectiveness of capturing both global and local structures in hyperbolic spaces.

The primary contributions of our work can be summarized as follows:
\vspace{-5pt}
\begin{itemize}
\item We propose a hyperbolic set-to-set distance measure that integrates the geodesic distance and topological distance for a comprehensive assessment of both global and local structures in hyperbolic sets.
\vspace{-5pt}
\item We introduce an effective and efficient manner to measure the topological distance between two hyperbolic sets, where we use a finite  Thue-Morse sequence for robust representations of topological structures of trees, while avoiding computation burdens. 
\end{itemize}
\section{Related Work}
\vspace{-5pt}
\subsection{Hyperbolic Geometry}
\vspace{-5pt}

Hyperbolic geometry, characterized by constant negative curvature, enables efficient embedding of hierarchical data as continuous analogs of trees. This property has driven research in hyperbolic learning across domains including few-shot learning~\citep{Anvekar_2023_CVPR, hamzaoui2024hyperbolic}, 3D vision~\citep{lin2023hyperbolic}, and image segmentation~\citep{chen2022hyperbolic}.
Early approaches to hyperbolic learning employed conventional neural networks for feature extraction in Euclidean space before projecting embeddings into hyperbolic space~\citep{khrulkov2020hyperbolic, atigh2022hyperbolic}. For instance, \cite{liu2020hyperbolic} extracted image features using convolutional networks in Euclidean space and mapped them to hyperbolic space via the exponential map, demonstrating improved robustness in zero-shot recognition tasks. However, these hybrid approaches often introduce data distortion due to the fundamental geometric incompatibility between their Euclidean backbones and hyperbolic target spaces. Recent work addresses this limitation by designing neural architectures that operate natively in hyperbolic spaces, including hyperbolic convolutional networks~\citep{bdeir2024fully} and hyperbolic residual networks~\citep{van2023poincare}, which preserve the geometric structure throughout the feature extraction process.

Recent work has explored hyperbolic geometry for metric learning, including point-to-set distances for few-shot classification~\citep{ma2022adaptive}, hyperbolic Vision Transformers~\citep{ermolov2022hyperbolic}, Busemann learning with ideal prototypes~\citep{ghadimi2021hyperbolic}, and uncertainty-aware embeddings for image retrieval~\citep{yan2023hyp} and segmentation~\citep{chen2023hyperbolic}. However, existing methods~\citep{dai2021hyperbolic, yan2021unsupervised} rely primarily on point-to-point distances that fail to capture complex hierarchical relationships between data clusters. Our work addresses this fundamental limitation by introducing novel set-to-set distance measures specifically formulated for hyperbolic geometry that preserve intrinsic curvature throughout similarity computation, enabling significantly better representation of hierarchical structures and outperforming existing techniques on complex nested hierarchies prevalent in real-world data.

\vspace{-5pt}
\subsection{Distance Metric Learning}
\vspace{-5pt}
Distance metric learning selects appropriate distance measures based on data structure and distributions, with widespread applications in image retrieval \citep{wu2021deep} and visual tracking \citep{wang2022negative}.
Traditional methods focus on point-to-point distance metrics (e.g., Mahalanobis distance) to evaluate similarity between individual data points \citep{Li_Li_Xie_Zhang_2022}. For computer vision tasks, point-to-set metrics have been developed to measure distance between an image and an image set \citep{wang2024spd}. For example, \cite{zhu2020progressive} introduced a progressive point-to-set metric learning model that improves few-shot classification through more accurate distance measurement between image sets and classes.

Set-to-set distance metrics \citep{qin2022spectral}, to which our method belongs, address tasks requiring measurement between two image sets. Ye et al. \citep{ye2020few} proposed neural networks for end-to-end learning of set-to-set distances. Unlike these approaches that operate in Euclidean space, our method leverages hyperbolic geometry to capture hierarchical structures in data sets, enabling set-to-set distance measurement that incorporates geometric information within the sets.

\vspace{-5pt}
\section{Preliminaries}
\vspace{-5pt}
\textbf{Notations.} In the following sections, $\mathbb{R}^n$ denotes $n$-dimensional Euclidean space and $\|\cdot\|$ denotes the Euclidean norm. 
The vectors are denoted by lower-case letters, such as $\boldsymbol{x}$ and $\boldsymbol{y}$. The matrices are denoted by upper-case letters, such as $\boldsymbol{M}$.

% \vspace{-5pt}
% \subsection{Lorentz Model}
% \vspace{-5pt}
% The Lorentz model $\mathbb{L}^n$ of an $n$-dimensional hyperbolic space is a manifold embedded in the $n + 1$ dimensional Minkowski space. The Lorentz model is defined as the upper sheet of a two-sheeted $n$-dimensional hyperbola with the metric $g^L$, which is
% \[
% \mathbb{L}^n = \{ x = (x_0, \ldots, x_n) \in \mathbb{R}^{n+1} : \langle x, x \rangle_L = -1, x_0 > 0 \},
% \]
% where the $\langle \cdot, \cdot \rangle_L$ represents the Lorentzian inner product.

% \noindent\textbf{Lorentzian inner product.}
% \begin{equation}
%     \langle x, y \rangle_L = x^T g^L y = -x_0 y_0 + \sum_{i=1}^n x_i y_i, \quad x, y \in \mathbb{R}^{n+1},
%     \label{eq:loentz_inner_product}
% \end{equation}
% where $g^L$ is a diagonal matrix with entries of 1s, except for the first element being $-1$. For any $x \in \mathbb{L}^n$, we can get that $x_0 = \sqrt{1 + \sum_{i=1}^n x_i^2}$. 

% \noindent\textbf{Lorentz distance function.}
% The distance in the Lorentz model is defined as

% \begin{equation}
%     d_L(x, y) = \operatorname{arcosh}(-\langle x, y \rangle_L).
%     \label{eq:lorentz_dis}
% \end{equation}

 \vspace{-5pt}
\subsection{Poincaré Model}
 \vspace{-5pt}
The Poincaré ball model of an $n$-dimensional hyperbolic space with curvature $c~(c<0)$ is defined as a Riemannian manifold $\left(\mathbb{B}^n_c, h^B_c\right)$, where $\mathbb{B}^n_c = \{ \boldsymbol{x}\in \mathbb{R}^n: -c\|\boldsymbol{x}\| < 1,c<0 \} $ is the open ball with radius $1/\sqrt{|c|}$ and $h^B_c$ is the Riemannian metric.
The tangent space at $\boldsymbol{x} \in \mathbb{B}^n_c$, a Euclidean space, is denoted by $T_{\boldsymbol{x}}\mathbb{B}^n_c$. 
We use the Möbius gyrovector space \citep{ungar2001hyperbolic} that provides operations for hyperbolic learning and several used operations are shown as follows. 

% More details regarding the Poincaré ball model and its properties are provided in the \textbf{Appendix} \ref{appendix:poincare}.

\noindent\textbf{Addition.} For a pair $\boldsymbol{x}, \boldsymbol{y} \in \mathbb{B}_c^n$, the Möbius addition is defined as
\begin{equation}
\boldsymbol{x} \oplus_c \boldsymbol{y}=\frac{\left(1-2 c\langle\boldsymbol{x}, \boldsymbol{y}\rangle-c\|\boldsymbol{y}\|^2\right) \boldsymbol{x}+\left(1+c\|\boldsymbol{x}\|^2\right) \boldsymbol{y}}{1-2 c\langle\boldsymbol{x}, \boldsymbol{y}\rangle+c^2\|\boldsymbol{x}\|^2\|\boldsymbol{y}\|^2}
.\end{equation}

\noindent\textbf{Hyperbolic $l_1$ distance.} The $l_1$ distance $d_c(\cdot, \cdot)$ between two points $\boldsymbol{x}, \boldsymbol{y} \in \mathbb{B}_c^n$ can be obtained as
\begin{equation}
d_c(\boldsymbol{x}, \boldsymbol{y})=\frac{2}{\sqrt{c}} \operatorname{arctanh}\left(\sqrt{c}\left\|-\boldsymbol{x} \oplus_c \boldsymbol{y}\right\|\right).
\label{eq:dist_0}
\end{equation} 

% \noindent\textbf{M\"{o}bius matrix multiplication.} In the Gyrovector space, the M\"{o}bius matrix multiplication $\otimes_c$ for the matrix $\boldsymbol{M} \in \mathbb{B}^{n\times n}$
% and vector $\boldsymbol{x} \in \mathbb{B}^{n}$ is defined as
% \begin{equation}
%     \boldsymbol{M} \otimes_c \boldsymbol{x} = \frac{1}{\sqrt{|c|}}\mathrm{tanh}(\frac{\|\boldsymbol{M}\boldsymbol{x}\|}{\|\boldsymbol{x}\|}\mathrm{arctanh}(\sqrt{|c|}\|\boldsymbol{x}\|))\frac{\boldsymbol{M}\boldsymbol{x}}{\|\boldsymbol{M}\boldsymbol{x}\|}.
%     \label{eq:bmm}
% \end{equation}

\noindent\textbf{Hyperbolic Einstein-Midpoint.} We use $\mathrm{Einstein}$ $\mathrm{mid}$-$\mathrm{point}$ as the counterpart of Euclidean averaging in hyperbolic space. The $\mathrm{Einstein}$ $\mathrm{mid}$-$\mathrm{point}$ has the most simple form in Klein model $\mathbb{K}$, thus for $\left(\boldsymbol{x}_1, \ldots, \boldsymbol{x}_N\right) \in \mathbb{B}$, we first map $\{\boldsymbol{x}\}$ from $\mathbb{B}$ to $\mathbb{K}$, then do the averaging in Klein model, and finally map the mean in $\mathbb{K}$ back to $\mathbb{B}$ to obtain the poincar\'e mean:
\begin{equation}
     \boldsymbol{u}_{i} =  \frac{2 \boldsymbol{x}_{i}}{1+c\left\|\boldsymbol{x}_{i}\right\|^{2}}, \ \
        \overline{\boldsymbol{u}} =  \frac{\sum_{i=1}^{N} \gamma_{i} \boldsymbol{u}_{i}}{\sum_{i=1}^{m} \gamma_{i}}, \ \
        \overline{\boldsymbol{x}} =  \frac{\overline{\boldsymbol{u}}}{1+\sqrt{1-c\|\overline{\boldsymbol{u}}\|^{2}}},
\label{eq:midpoint}
\end{equation}
where $\boldsymbol{u}_{i} \in \mathbb{K}$, $\overline{\boldsymbol{u}}$ is the mean in $\mathbb{K}$, $\overline{\boldsymbol{x}}$ is the mean in $\mathbb{B}$, and $\gamma_i=\frac{1}{\sqrt{1-c\left\|\boldsymbol{x}_i\right\|^2}}$ is the Lorentz factor. 

 \vspace{-7pt}
\subsection{Thue-Morse Sequence}
The Thue-Morse sequence is an optimal canonical sequence \cite{allouche2003automatic} for tree representation in spectral graph theory \cite{McKay}, minimizing redundancy through its overlap-free property. It is defined as follows:
\begin{equation}
    \boldsymbol{t}_0 = \boldsymbol{D},\quad \boldsymbol{t}_1 = \boldsymbol{D}\boldsymbol{A},\quad \boldsymbol{t}_2 = \boldsymbol{D}\boldsymbol{A}\boldsymbol{A}\boldsymbol{D},\quad \boldsymbol{t}_3 = \boldsymbol{DAADADDA},\quad \ldots,\quad \boldsymbol{t}_n = \boldsymbol{t}_{n-1} \boldsymbol{t}_{n-1}^*
    \label{eq:thue-morse}
\end{equation}
where $\boldsymbol{D}$ is the degree matrix, $\boldsymbol{A}$ is the adjacency matrix, $\boldsymbol{t}_{i-1}^{*}$ simply swaps $\boldsymbol{A}$ and $\boldsymbol{D}$ in $\boldsymbol{t}_{i-1}$. By $\boldsymbol{t}_n$ we mean $n$ terms in the Thue-Morse sequence as $\mathcal{T}_{n} = \{\boldsymbol{t}_i\}_{i=0}^n$.

% \zhi{conflict with 4.2.1. $\mathcal{T}_{n}$? $T_n(D,A)$ vs. $\boldsymbol{y}$, $D$ vs. $\boldsymbol{D}$ }

% We may use $T_n(D,A){T_n(D,A)}^T$ to simulate $p_jQp_j^T$, but since we are only concerned with spectra while the spectrum of $MM^T$ is just that of $M$ squared for any square matrix $M$ (with real spectrum), it suffices to consider the spectra of $T_n(D,A)$. Although the Thue-Morse sequence is uniformly recurrent, it is not eventually periodic, which suggests that it should avoid redundant computation of spectra of recurrent subwords. 

% \todo{Fullfill this section. checkout appendix B.[-1]}

 \vspace{-7pt}
\section{Hyperbolic Set-to-Set Distance}
 \vspace{-7pt}
% \todo{Full fill the problem formulation.}
% \todo{Add main theorem and theory results here.}
% \todo{Implementation Details.}

\subsection{Formulation}
 \vspace{-5pt}
\noindent\textbf{Point-to-Point Distance.} In hyperbolic space, similarity between data points is typically assessed by representing each data point as a feature point and computing their geodesic distance between each other,
% as illustrated in Eq.~\eqref{eq:point_distance}.
\begin{equation}
    d_c(\boldsymbol{x}, \boldsymbol{y})=\frac{2}{\sqrt{c}} \operatorname{arctanh}\left(\sqrt{c}\left\|-\boldsymbol{x} \oplus_c \boldsymbol{y}\right\|\right),
    \label{eq:point_distance}
\end{equation}
where $\boldsymbol{x}$ and $\boldsymbol{y}$ are hyperbolic features, $c$ is the curvature, $\oplus$ denotes the Möbius addition.

% \noindent\textbf{Set-to-Set Distance.} In many cases, data is more suitable to be represented as a set containing local features on the Euclidean space. Thus, set-to-set distances have been studied in Euclidean space.
% Given a representation $\mathbf{S}_{\boldsymbol{x}} = \{\boldsymbol{x}_{i}\}$, similarity is determined by averaging the pairwise similarities among all $\boldsymbol{x}_i$, which defines the distance measure.

% \begin{equation}
%     d_c(\mathbf{S}_{\boldsymbol{x}}, \mathbf{S}_{\boldsymbol{y}})= ||\mathbf{S}_{\boldsymbol{x}}  - \mathbf{S}_{\boldsymbol{y}}||.
%     \label{eq:set_distance}
% \end{equation}

\begin{figure}
    \centering
    \includegraphics[width=0.9\linewidth]{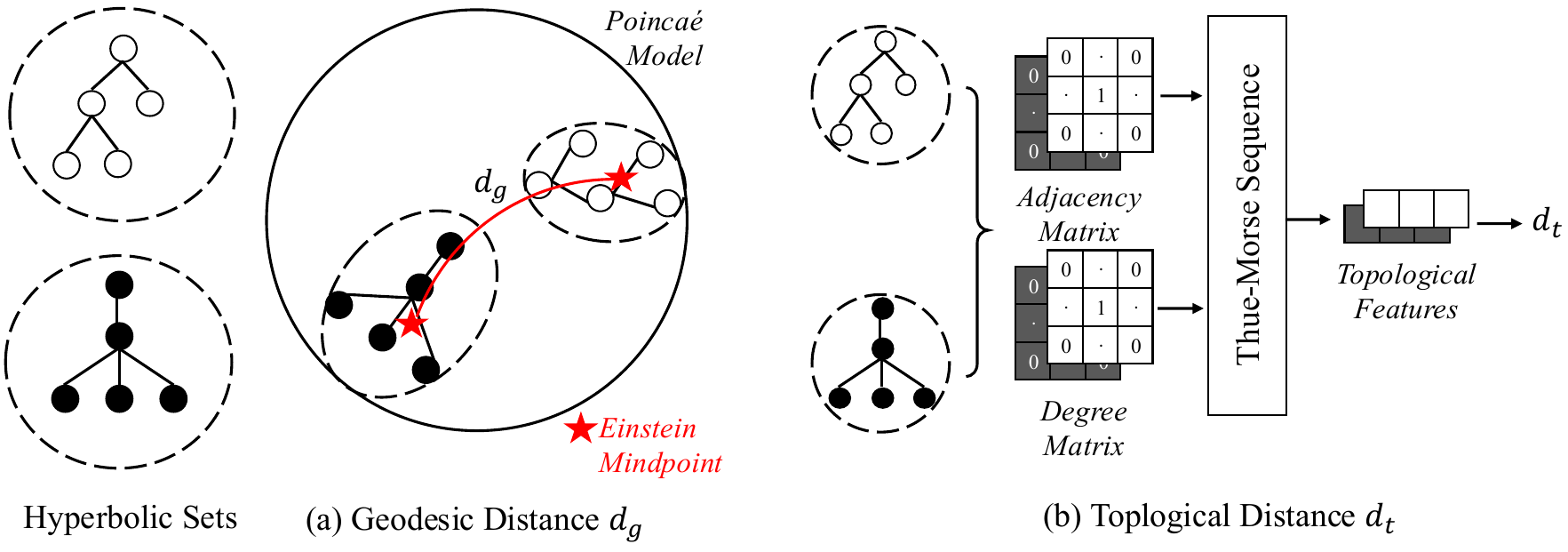}
    \caption{Pipeline of Hyperbolic Set-to-Set Distance Measure.}
    \label{fig:pipeline}
    \vspace{-10pt}
\end{figure}

\noindent\textbf{Our Approach.} We propose a set-to-set distance measure in hyperbolic space that utilizes both the global and local geometric structures within the set. Specifically, given two sets of hyperbolic representations $\mathbf{S}_{\boldsymbol{x}} = \{\boldsymbol{x}_{i}\}$ and $\mathbf{S}_{\boldsymbol{y}} = \{\boldsymbol{y}_{i}\}$, our distance measure is defined as follows:
\begin{equation}
    d_c(\mathbf{S}_{\boldsymbol{x}}, \mathbf{S}_{\boldsymbol{y}})= \lambda  d_{g} (\mathbf{S}_{\boldsymbol{x}}, \mathbf{S}_{\boldsymbol{y}}) + (1-\lambda )d_{t} (\mathbf{S}_{\boldsymbol{x}}, \mathbf{S}_{\boldsymbol{y}}) 
    \label{eq:ours_distance},
\end{equation}
where $d_g(\cdot, \cdot)$ is the \textbf{\textit{geodesic distance}} to measure the dissimilarity from a global perspective, and $d_t(\cdot,\cdot)$ is the  \textbf{\textit{topological distance}} to measure the similarity of the local geometrical structures of feature sets, and $\lambda$ is a hyper parameter to balance the two components. More details will be discussed in the following subsections. The pipeline is shown in \cref{fig:pipeline}.

 \vspace{-5pt}
\subsection{Topological Distance}
 \vspace{-5pt}

We model sets as trees in hyperbolic spaces and establish that the topology of a tree can be represented by the polynomial combinations of its adjacency and degree matrices. This representation is based on the following theorem.
 \vspace{-4pt}
\begin{theorem}\label{thm-McKay-main}
    Two trees $ T_\alpha $ and $ T_\beta $ are isomorphic if and only if for any polynomial $ p(\cdot,\cdot)$, $ p(\boldsymbol{A}_\alpha, \boldsymbol{D}_\alpha) $ and $ p(\boldsymbol{A}_\beta, \boldsymbol{D}_\beta) $ are cospectral
    , where $\boldsymbol{A}_{\alpha/\beta}$ and $\boldsymbol{D}_{\alpha/\beta}$ denote the adjacency and degree matrices of tree $T_{\alpha/\beta}$. We here use $\boldsymbol{A}$ and $\boldsymbol{D}$ to denote  adjacency matrices and degree matrices for simplicity.
     \vspace{-3pt}
    \begin{equation}
    % p(\boldsymbol{A}, \boldsymbol{D}) = \sum_{i=0}^\infty \sum_{j=0}^\infty c_{ij} \boldsymbol{A}^{i}\boldsymbol{D}^{j},
    p(\boldsymbol{A}, \boldsymbol{D}) = \left\{ \sum_{w}^{\infty}  \, w(\boldsymbol{A}, \boldsymbol{D})  \right\},
    \label{eq:poly}
\end{equation}
where $w(\boldsymbol{A},\boldsymbol{D})$ denotes any word in the alphabet $\{\boldsymbol{A},\boldsymbol{D}\}$.
\end{theorem}

 \vspace{-4pt}
\begin{remark}
~\Cref{thm-McKay-main} establishes that tree isomorphism can be determined through polynomial combinations $p(\boldsymbol{A}, \boldsymbol{D})$ of the adjacency matrix $\boldsymbol{A}$ (representing graph connectivity) and the degree matrix $\boldsymbol{D}$ (capturing local graph properties). 
Consequently,   the structure of each distinct tree $T$ corresponds to a unique polynomial combination $p(\boldsymbol{A}, \boldsymbol{D})$, providing a way to quantify topological similarities between trees. Detailed proof of ~\Cref{thm-McKay-main} is shown in the \Cref{app:proof}.
\end{remark}

While Theorem~\ref{thm-McKay-main} establishes a theoretical foundation, the infinite polynomial combination ~\cref{eq:poly} makes direct computation impossible. The following proposition addresses this challenge:

\begin{proposition}\label{prop-traces}
    For any two trees $T_\alpha$ and $T_\beta$, there exists a finite canonical sequence of words (\emph{i.e.}, $w(\boldsymbol{A}, \boldsymbol{D})$), such that linear combinations of these words span a dense subset of the polynomial space of $p(\boldsymbol{A}, \boldsymbol{D})$ described in Theorem \ref{thm-McKay-main}.

    This dense subset property of the canonical sequence implies that any polynomial combination $p(\boldsymbol{A}, \boldsymbol{D})$ can be approximated by linear combinations of these canonical words (\emph{i.e.},$w(\boldsymbol{A}, \boldsymbol{D})$ ). This approximation provides a practical way to represent tree structures using a finite canonical sequence, making the infinite polynomial space in ~\cref{eq:poly} computationally tractable.  

\end{proposition}

Next, we identify the Thue-Morse sequence as an efficient choice for these canonical words:

\vspace{-5pt}
\begin{proposition}\label{prop-TM-canonical}
The Thue-Morse sequence $\mathcal{T}_{n} = \{\boldsymbol{t}_i\}_{i=0}^n$ provides an optimal canonical sequence ~\citep{allouche2003automatic} of words for representing trees in the polynomial space of adjacency and degree matrices. Its optimality stems from its overlap-free property ~\citep{allouche2003automatic}, that is $\forall\,a\in\{\boldsymbol{A},\boldsymbol{D}\},\;\forall\,b \in \{\boldsymbol{A}, \boldsymbol{D}\} \setminus \{a\},\quad
a\,b\,a\,b\,a\;\notin\;\mathrm{Fact}(\mathcal{T}_n),$ where $\mathrm{Fact}(\cdot)$ is the set of all factors of $\mathcal{T}_n$.  The overlap-free property ensures each term $\boldsymbol{t}_i$ contributes unique structural information about the tree $T$ topology. %\zhi{use some mathematical symbols defined in 3.2}
% \zhi{that is, ... }.
\end{proposition}
\vspace{-5pt}
Furthermore, we establish a practical bound on the number (\emph{i.e.}, $n = 4$) of terms $\boldsymbol{t}_n$ needed: 

\vspace{-5pt}
% The optimality of the Thue-Morse sequence is particularly valuable for computational efficiency. Furthermore, we can establish a practical bound on the number of terms needed:
% \px{TODO explanation is needed}
\begin{corollary}\label{prop-TM-finite}
    According to Allouche et al \citep{allouche2003automatic}, the first four terms of the Thue-Morse sequence can represent a canonical sequence thus forming a basis for representing tree topology. This finite representation enables practical computation while maintaining theoretical guarantees of completeness. Detailed analysis is provided in the \Cref{app:tm-analysis}.
\end{corollary}
\vspace{-5pt}
% \begin{corollary}\label{prop-TM-finite}
%     For practical applications in tree comparison, the first four terms of the Thue-Morse sequence ($\boldsymbol{t}_0=\boldsymbol{D}$, $\boldsymbol{t}_1=\boldsymbol{DA}$, $\boldsymbol{t}_2=\boldsymbol{DAAD}$, $\boldsymbol{t}_3=\boldsymbol{DAADADDA}$) are sufficient to form an effective canonical sequence \px{show the key idea why this is sufficient? details in supp} \zhi{the scheme to show this?}. These terms capture the essential topological information while maintaining minimal redundancy due to their overlap-free property.
% \end{corollary}

\vspace{-5pt}
\subsubsection{Implementation} %\px{make them consistent}
\vspace{-5pt}
% \noindent\textbf{Graph Construction.}
For each set $\mathbf{S}_{\boldsymbol{x}} = \{\boldsymbol{x}\}$, we construct its graph representation by first calculating pairwise hyperbolic $l_1$ distances between feature points using~\cref{eq:dist_0}. These distances form the adjacency matrix $\boldsymbol{A}_{\boldsymbol{x}}$, from which we derive the corresponding degree matrix $\boldsymbol{D}_{\boldsymbol{x}}$. For simplicity, we denote these matrices as $\boldsymbol{A}$ and $\boldsymbol{D}$ in subsequent discussions.

% \noindent\textbf{Thue-Morse Sequence for Tree Representation.} 
Based on Theorem~\ref{thm-McKay-main} and Proposition~\ref{prop-TM-canonical}, we use the Thue-Morse sequence to generate canonical words that efficiently represent tree topology. Following Corollary~\ref{prop-TM-finite}, we use the first $n$ terms of this sequence to create a compact representation of tree topology.

\noindent\textbf{Topological Distance Measure.}
To quantify the topological difference between two sets $\mathbf{S}_{\boldsymbol{x}} =\{\boldsymbol{x}_{i}\}$ and $\mathbf{S}_{\boldsymbol{y}} = \{\boldsymbol{y}_{i}\}$, we compute their respective Thue-Morse sequences $\mathcal{T}^{x}_{n=4} = \{\boldsymbol{t}^{x}_{i}\}_{i=0}^{n=4}$ and $\mathcal{T}^{y}_{n=4} = \{\boldsymbol{t}^{y}_{i}\}_{i=0}^{n=4}$ using the adjacency and degree matrices derived from each set. The topological distance is then defined as the average Frobenius norm of the differences between corresponding sequence elements:
\begin{equation}
    d_{t}(\mathbf{S}_{\boldsymbol{x}}, \mathbf{S}_{\boldsymbol{y}}) = \frac{1}{n+1} \sum_{i=0}^{n=4}||\boldsymbol{t}_{i}^{x} - \boldsymbol{t}_{i}^{y}||_{F}.
\end{equation}
This measure effectively captures structural differences between the sets while maintaining computational efficiency.

% \subsection{Geodesic Distance}
% The geodesic distance is defined as the distance $ d_g(\cdot, \cdot) $ between the central points of two sets. For two sets 
% $\mathbf{S}_{\boldsymbol{x}} = \{\boldsymbol{x}_i\}, \mathbf{S}_{\boldsymbol{y}} = \{\boldsymbol{y}_{i}\}$, we first compute the $\mathrm{Einstein}-$midpoints $\overline{\boldsymbol{x}}$, $\overline{\boldsymbol{y}}$ of the two sets via~\cref{eq:midpoint}. The geodesic distance is defined as 
% \begin{equation}
%     d_{g}(\mathbf{S}_{\boldsymbol{x}}, \mathbf{S}_{\boldsymbol{y}}) = 
%     \frac{2}{\sqrt{c}} \operatorname{arctanh}\left(\sqrt{c}\left\|-\overline{\boldsymbol{x}} \oplus_c \overline{\boldsymbol{y}}\right\|\right).
% \end{equation}
\vspace{-5pt}
\subsection{Geodesic Distance}  
\vspace{-5pt}
The geodesic distance measures the global positional relationship between sets in hyperbolic space by computing the distance between their central points. For sets $\mathbf{S}_{\boldsymbol{x}} = \{\boldsymbol{x}_i\}$ and $\mathbf{S}_{\boldsymbol{y}} = \{\boldsymbol{y}_i\}$, we first calculate their Einstein midpoints $\overline{\boldsymbol{x}}$ and $\overline
{\boldsymbol{y}}$ using~\cref{eq:midpoint}, which aggregates information from all elements in each respective set. 
 The geodesic distance between sets is then formulated as:

\begin{equation}
    d_{g}(\mathbf{S}_{\boldsymbol{x}}, \mathbf{S}_{\boldsymbol{y}}) = 
    \frac{2}{\sqrt{c}} \operatorname{arctanh}\left(\sqrt{c}\left\|-\overline{\boldsymbol{x}} \oplus_c \overline{\boldsymbol{y}}\right\|\right),
\end{equation}
 
where $c$ represents the curvature of the hyperbolic space and $\oplus_c$ denotes the Möbius addition operation. This formulation precisely measures the distance of the geodesic path connecting the two set representatives.

 \vspace{-10pt}
\subsection{$\lambda$ Adapter ($\mathcal{G}$)}
 \vspace{-5pt}
We introduce a $\lambda$ adapter that dynamically balances topological and geodesic distances based on the properties of the input sets. For sets $\{\boldsymbol{x}\}_n$ and $\{\boldsymbol{y}\}_n$, our approach first captures their interactions using a two-layer cross-attention module $\mathrm{Att}(\cdot,\cdot)$. The output from this attention mechanism is then processed by a subnetwork $g(\cdot)$, which consists of two linear layers followed by a softmax function, to produce the weighting parameter $\lambda$, as shown in~\cref{eq:lambda_adapter}
\begin{equation}
    \lambda = \mathcal{G}(\{\boldsymbol{x}\}_n,\{\boldsymbol{y}\}_n) =\mathrm{softmax}(g(\mathrm{Att}(\{\boldsymbol{x}\}_n,\{\boldsymbol{y}\}_n)))
    \label{eq:lambda_adapter}
\end{equation}
\subsection{Training process}
 \vspace{-5pt}
    The training process of HS2SD is summarized in Algorithm \ref{al:training}. We sample support set $\mathcal{D}_{s}$ and query set $\mathcal{D}_{q}$ from the training data, then extract features using the feature extractor $\mathcal{E}$. We calculate Einstein midpoints for hyperbolic feature sets and generate the adaptive parameter $\lambda$ using our $\lambda$ adapter $\mathcal{G}$. The final distance is computed by combining topological and geodesic distances according to Eq.~(\ref{eq:ours_distance}), which is then used to calculate the cross-entropy loss for model optimization.

\begin{figure}[!ht]
% \vspace{-10pt}
\begin{minipage}[c]{0.62\linewidth}
  \vspace{0pt} % 确保顶部对齐
  \centering
  \footnotesize
  \captionsetup{type=algorithm, labelfont=bf, skip=2pt}
  \begin{algorithm}[H]
    \caption{Training Process of \textbf{HS2SD}}\label{al:training}
    \begin{algorithmic}[1]
      \REQUIRE Training set $\mathcal{D}$
      \ENSURE Updated feature extractor $\mathcal{E}$ and lambda adapter $\mathcal{G}$
      \WHILE{Not converged}
          \STATE Sample support set $\mathcal{D}_s$ and query set $\mathcal{D}_q$ from $\mathcal{D}$
          \STATE Extract features $F_s = \mathcal{E}(\mathcal{D}_s)$, $F_d = \mathcal{E}(\mathcal{D}_q)$
          \STATE Compute Einstein mid-point and get matrix $\boldsymbol{W}$ 
          \STATE Generate pair-wise $\lambda$ using $\mathcal{G}$ via Eq.(~\ref{eq:lambda_adapter})
          \STATE Compute the distance as logits via Eq.(~\ref{eq:ours_distance})
          \STATE Compute cross-entropy loss and update $\mathcal{E}$ and $\mathcal{G}$
      \ENDWHILE
    \end{algorithmic}
  \end{algorithm}
\end{minipage}
\hfill
\begin{minipage}[c]{0.38\linewidth}
  \vspace{-6pt}
  \centering
  \footnotesize
  \captionsetup{type=table, labelfont=bf, skip=2pt}
  \captionof{table}{Statistics of EM Datasets}\label{tab:dataset_statistics}
  \vspace{6pt} % 标题与表格间距
  \resizebox{\linewidth}{!}{
    \renewcommand{\arraystretch}{1.2} % 增加行高
    % \begin{tabular}{@{}lcccc@{}}
    %   \toprule
    %   \textbf{Dataset} & \textbf{\#Tuples} & \textbf{\#Cand} & \textbf{\#Match} & $\delta_{\text{H}}$ \\
    %   \midrule
    %   DA & 2,471-2,260 & 5.58M & 2,220 & 0.333 \\
    %   AG & 1,295-2,150 & 2.78M & 1,167 & 0.282 \\
    %   WA & 1,688-5,249 & 8.86M & 962 & 0.329 \\
    %   DS & 2,576-10.7K & 27.5M & 5,347 & 0.302 \\
    %   FZ & 293-238 & 69K & 110 & 0.303 \\
    %   DA-d & 2,471-2,260 & 5.58M & 2,220 & 0.309 \\
    %   DS-d & 2,576-10.7K & 27.5M & 5,347 & 0.304 \\
    %   \bottomrule
    % \end{tabular}
    \begin{tabular}{@{}lcccc@{}}
      \toprule
      \textbf{Dataset} & \textbf{\#Tuples} & \textbf{\#Match} & $\delta$-\textbf{Hyp.} \\
      \midrule
      DA & 2,471-2,260 & 2,220 & 0.333 \\
      AG & 1,295-2,150 & 1,167 & 0.282 \\
      WA & 1,688-5,249 & 962 & 0.329 \\
      DS & 2,576-10694 & 5,347 & 0.302 \\
      FZ & 293-238 & 110 & 0.303 \\
      DA-d & 2,471-2,260 & 2,220 & 0.309 \\
      DS-d & 2,576-10694 & 5,347 & 0.304 \\
      \bottomrule
    \end{tabular}
  }
\end{minipage}
\vspace{-10pt}
\end{figure}
 
% \subsection{Implementation}
% \subsubsection{Graph Construction}
% We construct an adjacency matrix by calculating the pairwise $l_1$ distances between each hyperbolic feature point. Based on this adjacency matrix, we further derive the degree matrix to facilitate subsequent analyses and computations.

% \noindent\textbf{Distance Measure.}

% \newpage
\section{Experiment} \label{sec:experiment}

\subsection{Unsupervised Entity Matching}
\subsubsection{Experimental Setup}

Unsupervised entity matching (EM) is a challenging task that aims to identify whether different entity descriptions correspond to the same real-world entity without relying on labeled training data.

\noindent\textbf{Dataset.}
We conduct experiments on 7 widely-used EM datasets~\citep{mudgal2018deep}, including DBLP-ACM (DA), Amazon-Google (AG), Walmart-Amazon (WA), DBLP-Scholar (DS), Fodors-Zagats (FZ), DBLP-ACM-dirty (DA-d), DBLP-Scholar-dirty (DS-d) and Walmart-Amazon-dirty (WA-d). Following CampER, we filter out tuples that are not included in the ground truth for fair comparison. Each dataset is structured into two tables, with each table comprising multiple tuples. The objective is to predict the matched tuple pairs between the two tables. The statistics of the datasets are summarized in Table \ref{tab:dataset_statistics}.

\noindent$\delta-$\textbf{hyperbolicity.}
% \px{CVPR 907-910\\}
% \px{1. Hyperbolicity is to measure the hierarchical level in data. 2. property of delta: lower indicates higher hierarchy 3. We measure the hyperbolicity of the features of all the datasets extracted via our feature extractor. 4. delta relatively low, indecates our motivation to test on these datasets.}
$\delta-$Hyperbolicity quantifies the hierarchical structure of data, with lower $\delta$ values indicating stronger hierarchy \citep{gromov1987hyperbolic}. We compute the $\delta$-hyperbolicity of features extracted by our model across the 7 EM datasets, as shown in Table \ref{tab:dataset_statistics}. The consistently low $\delta$ values demonstrate the datasets exhibit inherent hierarchical structure, justifying their suitability for our experiments.

\noindent\textbf{Metric.}
% \px{List the eval metric that we use in the following experiments. For us, we use Precision, Recall, and F1 score. Make sure you include the equation of the metrics. Besides, add short rationale about why we use these metrics (also can check the baseline). }
% \ww{1.Only F1 is used for evaluation. 2.Camper did not elaborate on why the f1 score was used. I also searched for two EM articles but could not find it.}
% \px{I see. Then only describe the F1 score here. you can follow camper's expression.}
Following most EM works ~\citep{guo2023camper, tu2022domain}, we evaluate the performance of EM results using the F1-score, which is defined as the harmonic mean of precision ($pre$) and recall ($rec$), computed as $
\mathrm{F1} = \frac{2(pre \times rec)}{(pre + rec)}.$

\noindent\textbf{Training Details.}
% \px{Details about the training setup like epoches, backbone, hyper parameter like learning rate, optimizer, etc.}
% Following~\cite{guo2023camper}, we adopt RoBERTa~\cite{liu2019roberta} as our backbone model, and the dimension of output vector is 768. We use Adam ~\cite{kingma2014adam} for optimization.
% (to be updated)
% We choose Camper ~\citep{guo2023camper} as our baseline. 
% Our training strategy largely follows CampER ~\citep{guo2023camper}. 
% 
We use RoBERTa~\citep{liu2019roberta} as our feature extractor, projecting the features onto the Lorentz model and measuring similarity with our HS2SD approach. For privacy preservation, we train two independent models using contrastive learning, exchanging parameters only after each epoch. The hyperbolic space curvature $c$ ranges from 0.005 to 0.2. We use Adam optimizer with learning rate 3e-5 and train for 20 epochs across all datasets.

\vspace{-5pt}
\begin{table*}[htbp]
    \caption{Results on EM datasets. All F1 scores are reported as percentages. 
    {\color{red}$\uparrow$} indicates improvements over the previous best method (CampER), and gray numbers denote absolute differences.}
    \label{tab:em_results}
    \centering
    \resizebox{0.85\textwidth}{!}{
    \begin{tabular}{lcccccccc}
        \toprule
        \textbf{Method}  & \textbf{DA} & \textbf{AG} & \textbf{WA} & \textbf{DS} & \textbf{FZ} & \textbf{DA-d} & \textbf{DS-d} & \textbf{Avg} \\
        \midrule
        BF ~\citep{schnell2009privacy} & 84.6 & 47.5 & 31.0 & - & 91.1 & 73.4 & - & 65.5 \\
        ZeroER ~\citep{wu2020zeroer}         & 98.2 & 43.5 & 67.2 & 92.7 & 95.5 & 60.7 & 51.1 & 72.7 \\
        EmbDI   ~\citep{cappuzzo2020creating}        & 98.9 & 71.2 & 74.7 & 94.9 & 99.1 & 98.5 & 94.8 & 90.3 \\
        Auto-FuzzyJoin     ~\citep{li2021auto}    & 96.8 & 43.4 & 65.5 & 85.8 & 85.7 & 89.8 & 78.0 & 77.9 \\
        CampER ~\citep{guo2023camper}         & \textbf{99.0} & 73.4 & 83.2 & 95.0 & 97.7 & \textbf{98.7} & 95.2 & 91.7 \\
        \midrule
        HS2SD (ours)    
        & \textbf{99.0} {\scriptsize\textcolor{gray}{0.0}} 
        & \textbf{79.2} {\color{red}$\uparrow$}{\scriptsize\textcolor{gray}{5.8}} 
        & \textbf{86.3} {\color{red}$\uparrow$}{\scriptsize\textcolor{gray}{3.1}} 
        & \textbf{96.4} {\color{red}$\uparrow$}{\scriptsize\textcolor{gray}{1.4}} 
        & \textbf{100.0} {\color{red}$\uparrow$}{\scriptsize\textcolor{gray}{2.3}} 
        & 98.5 {\scriptsize\textcolor{gray}{-0.2}} 
        & \textbf{96.3} {\color{red}$\uparrow$}{\scriptsize\textcolor{gray}{1.1}} 
        & \textbf{93.7} {\color{red}$\uparrow$}{\scriptsize\textcolor{gray}{2.0}} \\
        \bottomrule
    \end{tabular}
    }
\end{table*}
\vspace{-5pt}
% \begin{table*}[htbp]
%     \caption{Results on EM datasets. All F1 scores are reported as percentages.}
%     \label{tab:em_results}
%     \centering
%     \resizebox{0.75\textwidth}{!}{
%     \begin{tabular}{lcccccccc}
%         \toprule
%         \textbf{Method}  & \textbf{DA} & \textbf{AG} & \textbf{WA} & \textbf{DS} & \textbf{FZ} & \textbf{DA-d} & \textbf{DS-d} & \textbf{Avg} \\
%         \midrule
%         BF ~\citep{schnell2009privacy} & 84.6 & 47.5 & 31.0 & - & 91.1 & 73.4 & - & 65.5 \\
%         ZeroER ~\citep{wu2020zeroer}         & 98.2 & 43.5 & 67.2 & 92.7 & 95.5 & 60.7 & 51.1 & 72.7 \\
%         EmbDI   ~\citep{cappuzzo2020creating}        & 98.9 & 71.2 & 74.7 & 94.9 & 99.1 & 98.5 & 94.8 & 90.3 \\
%         Auto-FuzzyJoin     ~\citep{li2021auto}    & 96.8 & 43.4 & 65.5 & 85.8 & 85.7 & 89.8 & 78.0 & 77.9 \\
%         CampER ~\citep{guo2023camper}         & \textbf{99.0} & 73.4 & 83.2 & 95.0 & 97.7 & \textbf{98.7} & 95.2 & 91.7 \\
%         \midrule
%         HS2SD (ours)    & \textbf{99.0} & \textbf{79.2} & \textbf{86.3} & \textbf{96.4} & \textbf{100.0} & 98.5 & \textbf{96.3} & \textbf{93.7} \\
%         \bottomrule
%     \end{tabular}
%     }
    
% \end{table*}
% \vspace{-5pt}
% \px{\textbf{NOTE}: when you using citation  Use ~\cite{guo2023camper} or CampER ~\citep{guo2023camper}}

\subsubsection{Experimental Findings}
\vspace{-5pt}
% \px{Brief intro of the experiment setting + Table + Result analysis}
% \px{Table : baseline + similar methods + ours}
% \px{I suggest we abandon the saying like "CampER-HS2SD". You may directly use "HS2SD (Ours) or Ours"}

We compare our method with five unsupervised EM baselines, including two privacy-aware methods BF~\citep{schnell2009privacy} and CampER~\citep{guo2023camper} and three non-private alternatives: ZeroER~\citep{wu2020zeroer}, Auto-FuzzyJoin~\citep{li2021auto}, and EmbDI~\citep{cappuzzo2020creating}. Table \ref{tab:em_results} presents the performance comparison across 7 EM datasets. Our method demonstrates significant advantages on complex datasets with ambiguous matching candidates, outperforming CampER (2nd highest) by $\uparrow$\textbf{5.7\%} on AG and $\uparrow$\textbf{3.1\%} on WA. These results confirm that HS2SD effectively captures both global and local information, enabling superior handling of challenging datasets. Across all datasets, our method achieves the highest average F1 score, demonstrating its overall effectiveness and adaptability.

\subsection{Standard Image Classification}
\subsubsection{Experimental Setup}
\noindent\textbf{Dataset.} We conduct experiments on three popular datasets, namely, MNIST\cite{mnist}, CIFAR10\cite{cifar}, and CIFAR100\cite{cifar}.  More details regarding the datasets are provided in the \textbf{Appendix} \ref{appendix:exp}.

\noindent\textbf{Training Details.} We use different backbone networks for each dataset, aligning with previous work to have a fair comparison. For MNIST, we use a LeNet-like architecture\citep{lecun1998gradient}. For CIFAR10 and CIFAR100, we use Wide-ResNet $28\times 2$\citep{zagoruyko2016wide} as the backbone network. We train our model for 100 epochs with the Adam optimizer. The learning rate is set to $0.001$ at first and decayed every 15 epochs with a decay rate of $0.5$.

\begin{wraptable}{r}{0.4\textwidth}
\vspace{-40pt}
\centering
\caption{Classification accuracy (\%)}\label{tab:results_standard}
\adjustbox{max width=\linewidth,valign=t}{ % 垂直顶部对齐
\begin{tabular}{@{}lccc@{}}
\toprule
\textbf{Method} & \textbf{MNIST} & \textbf{CIFAR10} & \textbf{CIFAR100} \\ 
\midrule
Hyp-Optim & 94.42 & 88.82 & 72.26 \\
HNN++ & 95.01 & 91.22 & 73.65 \\
Hyp-ProtoNet & 93.53 & 93.30 & 73.83 \\
HS2SD & \textbf{98.92} & \textbf{93.60} & \textbf{75.43} \\ 
\bottomrule
\end{tabular}
}
\vspace{-10pt}
\end{wraptable}

\subsubsection{Experimental Findings}
 We evaluate HS2SD against hyperbolic learning approaches including Hyp-Optim\citep{ganea2018hyperbolic}, HNN++\citep{shimizu2020hyperbolic}, and Hyp-ProtoNet\citep{khrulkov2020hyperbolic}. Table \ref{tab:results_standard} presents the comparative results across standard image classification benchmarks. The results indicate that HS2SD performs favorably across all datasets, with particularly notable performance on MNIST and the more challenging CIFAR100 dataset. These results suggest that our approach effectively leverages hyperbolic geometry to represent the hierarchical relationships present in visual data.

% \begin{table*}[htbp]
% \caption{Results on standard image classification datasets}\label{tab:results_standard}
% \centering
% \resizebox{0.5\textwidth}{!}{ 
% \begin{tabular}{lccc}
% \toprule
% \textbf{Method} & \textbf{MNIST} & \textbf{CIFAR10} & \textbf{CIFAR100} \\ \midrule
% Hyp-Optim \citep{ganea2018hyperbolic} & 94.42 & 88.82 & 72.26 \\
% HNN++ \citep{shimizu2020hyperbolic} & 95.01 & 91.22 & 73.65 \\
% Hyp-ProtoNet \citep{khrulkov2020hyperbolic} & 93.53 & 93.30 & 73.83 \\
% HS2SD & \textbf{98.92} & \textbf{93.60} & \textbf{75.43} \\ \bottomrule
% \end{tabular}
% }
% \end{table*}

\subsection{Few-shot Image Classification}
% \subsubsection{Experiment Setup}
\subsubsection{Experimental Setup}
\noindent\textbf{Dataset.}
% \px{Feel free to check this link \href{https://www.overleaf.com/8117958577ffbgthcjfswj#dbb429}{CVPR25} Sec5.2 and Appendix F.3 to find the information that you can directly copy into this section.}
We conduct experiments on three popular few-shot learning datasets, namely, miniImagenet ~\citep{vinyals2016matching}, tieredImagenet ~\citep{ren2018meta} and CIFAR-FewShot (CIFAR-FS) ~\citep{bertinetto2018meta}. Full details of the datasets and implementation details are provided in Appendix \ref{appendix:exp}. These datasets have been validated in prior work ~\citep{khrulkov2020hyperbolic} as suitable for modeling in hyperbolic space for the few-shot image classification tasks.

\noindent\textbf{Training Details.}
% \px{Do not use this writing style ,you can just say 'Following \citet{khrulkov2020hyperbolic}, we xxx'}
Following Hyp-ProtoNet\cite{khrulkov2020hyperbolic}, we use ResNet12 as our backbone, aligning with the existing methods to ensure fairness. We remove the global average pooling layer to generate a feature set for each image. Specifically, the model encodes an image of size 84 × 84 into a feature map of size $640 \times 5 \times 5$, which is subsequently expanded along the spatial dimensions to produce 25 features, each of dimension 640. Hyper-parameters, such as learning rate and temperature, are selected through a grid search on the validation set. We train the model for 100 epochs and select the model that performs best on the validation set for evaluation.

% \noindent\textbf{Evaluation Metric.}
% \ww{We use accuracy as the evaluation metric, is there a need to write something about this?}

% 第一个表格（miniImagenet和CUB）
% \begin{table}[t]
% \caption{Experimental results on miniImagenet and CUB datasets.}
% \label{tab:results1}
% \centering
% \begin{tabular}{@{}lrrrrrr@{}}
% \toprule
% \textbf{Method} & \multicolumn{2}{c}{miniImagenet} & \multicolumn{2}{c}{CUB} \\
% \cmidrule(lr){2-3} \cmidrule(lr){4-5}
%  & 1-shot & 5-shot & 1-shot & 5-shot \\ 
% \midrule
% Hyp-ProtoNet & 59.47 ± 0.20 & 76.84 ± 0.14 & 64.02 ± 0.24 & 82.53 ± 0.14 \\
% Hyp-Kernel & 61.04 ± 0.21 & 77.33 ± 0.15 & 70.98 ± 0.22 & 85.21 ± 0.13 \\
% HS2SD (ours) & \textbf{64.37 ± 0.21} & \textbf{80.88 ± 0.15} & \textbf{67.97 ± 0.24} & -- ± -- \\
% \bottomrule
% \end{tabular}
% \end{table}

% % 第二个表格（tieredImagenet和CIFAR-FS）
% \begin{table}[t]
% \caption{Experimental results on tieredImagenet and CIFAR-FS datasets.}
% \label{tab:results2}
% \centering
% \begin{tabular}{@{}lrrrrrr@{}}
% \toprule
% \textbf{Method} & \multicolumn{2}{c}{tieredImagenet} & \multicolumn{2}{c}{CIFAR-FS} \\
% \cmidrule(lr){2-3} \cmidrule(lr){4-5}
%  & 1-shot & 5-shot & 1-shot & 5-shot \\ 
% \midrule
% Hyp-ProtoNet & -- ± -- & -- ± -- & -- ± -- & -- ± -- \\
% Hyp-Kernel & 57.78 ± 0.23 & 76.48 ± 0.18 & -- ± -- & -- ± -- \\
% HS2SD (ours) & \textbf{71.22 ± 0.23} & \textbf{86.03 ± 0.15} & \textbf{73.40 ± 0.21} & \textbf{86.51 ± 0.15} \\
% \bottomrule
% \end{tabular}
% \end{table}

% Please add the following required packages to your document preamble:
\vspace{-7pt}
\begin{table*}[htbp]
  \caption{Results on few-shot classification datasets.‘*’
denotes results reproduced via official codebases.}\label{tab:fs_results_all}
  \centering
  \resizebox{\textwidth}{!}{ 
  \begin{tabular}{lccccccc}
    \toprule
    \multirow{2}{*}{\textbf{Method}} &\multirow{2}{*}{\textbf{Space}} & \multicolumn{2}{c}{\textbf{miniImagenet}} & \multicolumn{2}{c}{\textbf{tieredImagenet}} & \multicolumn{2}{c}{\textbf{CIFAR-FS}} \\
    & & 1-shot & 5-shot & 1-shot & 5-shot & 1-shot & 5-shot \\ 
    \midrule
    
    TADAM\citep{oreshkin2018tadam} &Euc & 58.50 ± 0.30 & 76.70 ± 0.30 & -- ± -- & -- ± -- & -- ± -- & -- ± -- \\
    ProtoNet\citep{snell2017prototypical} &Euc & 60.37 ± 0.83 & 78.02 ± 0.57 & 65.65 ± 0.92 & 83.40 ± 0.65 & 72.20 ± 0.70 & 83.50 ± 0.50 \\
    MetaOptNet\citep{lee2019meta} &Euc & 62.64 ± 0.82 & 78.63 ± 0.46 & 65.99 ± 0.72 & 81.56 ± 0.53 & 72.60 ± 0.70 & 84.30 ± 0.50 \\ \midrule
    C-HNN\cite{guo2022clipped} &Hyp & 53.01 ± 0.22 & 72.66 ± 0.15 & -- ± -- & -- ± -- & -- ± -- & -- ± -- \\
    Hyp-Kernel\citep{fang2021kernel} &Hyp & 61.04 ± 0.21 & 77.33 ± 0.15 & 57.78 ± 0.23 & 76.48 ± 0.18 & -- ± -- & -- ± -- \\
    Hyp-ProtoNet\citep{khrulkov2020hyperbolic} &Hyp & 59.47 ± 0.20 & 76.84 ± 0.14 & 59.00 ± 0.23* & 73.98 ± 0.19* & 65.43 ± 0.23* & 78.59 ± 0.17* \\ \midrule
    HS2SD (ours) &Hyp & \textbf{64.37 ± 0.21} & \textbf{80.88 ± 0.14} & \textbf{71.22 ± 0.23} & \textbf{86.03 ± 0.15} & \textbf{73.63 ± 0.21} & \textbf{86.51 ± 0.15} \\ 
    \bottomrule
  \end{tabular}
  }
\end{table*}
\subsubsection{Experimental Findings}
We compare our method with three hyperbolic methods, including Hyp-ProtoNet\citep{khrulkov2020hyperbolic}, Hyp-Kernel\citep{fang2021kernel} and C-HNN\citep{guo2022clipped}, as shown in Table \ref{tab:fs_results_all}. Our method is most  similar to Hyp-ProtoNet, which performs prototype alignment in hyperbolic space. However, by leveraging local structural information of data points, our method demonstrates substantial performance gains across both 1-shot and 5-shot scenarios. Specifically, our method outperforms Hyp-ProtoNet by 4.90\% and 4.04\% on miniImagenet, and achieves even larger improvements of 12.22\% and 12.05\% on tieredImagenet in 1-shot and 5-shot settings, respectively. Compared to other hyperbolic methods, our approach shows clear advantages, surpassing Hyp-Kernel by 3.33\% in 1-shot and 3.55\% in 5-shot scenarios on miniImagenet, while significantly outperforming C-HNN by 11.36\% and 8.22\% in corresponding settings. We also compare our method with 3 popular Euclidean methods: TADAM\citep{oreshkin2018tadam}, ProtoNet\citep{snell2017prototypical}, and MetaOptNet\citep{lee2019meta}. Our approach consistently outperforms these Euclidean baselines across all datasets.

% TADAM\cite{oreshkin2018tadam},  ProtoNet\cite{snell2017prototypical},  MetaOptNet\citep{lee2019meta}.

% As shown in Table~\ref{tab:fs_results_all}, our method outperforms the baseline on all datasets under the one-shot setting with obvious improvement (1\% in average), demonstrating the effectiveness of the detailed patterns provided by our HS2SD measurement. Our method is comparable to baseline under 5 shot setting with a small improvement(0.25\% in average).We believe this is due to the fact that averaging all shots can distort the geometric structure, leading to reduced effectiveness of our geometry distance measurement.
% \px{Analysis according to the dataset. }

\vspace{-10pt}
\subsection{Ablation}
% \px{All the ablation experiments is to show the effectiveness of several module. It should be format as 'Main claim/observation/insight' + result + analysis/discussion.}
We conduct ablation studies on three few-shot image classification datasets to evaluate the effectiveness of the geodesic distance measure($d_{g}$), topological distance ($d_{t}$), and the lambda adapter($\mathcal{G}$). We also explore the impact of curvature $c$ and balancing parameter $\lambda$ on EM tasks. The results are shown in Table~\ref{tab:fs_ablation} and Fig.~\ref{fig:ablation_lambdac}.

\vspace{-10pt}
\begin{table*}[htbp]
  \caption{Ablation studies on few-shot image classification tasks.}\label{tab:fs_ablation}
  \centering
  \resizebox{0.8\textwidth}{!}{ 
  \begin{tabular}{lcccccc}
    \toprule
    \multirow{2}{*}{\textbf{Method}}  & \multicolumn{2}{c}{\textbf{miniImagenet}} & \multicolumn{2}{c}{\textbf{tieredImagenet}} & \multicolumn{2}{c}{\textbf{CIFAR-FS}} \\
     & 1-shot & 5-shot & 1-shot & 5-shot & 1-shot & 5-shot \\ 
    \midrule
    \textbf{Ours w/o $d_{g}$} & 61.07 $\pm$ 0.20 & 76.18 $\pm$ 0.16 & 67.78 $\pm$ 0.24 & 81.93 $\pm$ 0.27 & 71.03 $\pm$ 0.22 & 83.22 $\pm$ 0.16 \\
\textbf{Ours w/o $d_{t}$} & 63.00 $\pm$ 0.20 & 80.49 $\pm$ 0.14 & 70.87 $\pm$ 0.23 & 85.90 $\pm$ 0.15 & 73.28 $\pm$ 0.21 & 86.34 $\pm$ 0.15 \\
\textbf{Ours w/o $\mathcal{G}$} & 63.31 $\pm$ 0.20 & 79.66 $\pm$ 0.15 & 69.74 $\pm$ 0.23 & 84.12 $\pm$ 0.16 & 72.49 $\pm$ 0.22 & 85.32 $\pm$ 0.15 \\
\textbf{Ours} & \textbf{64.37 $\pm$ 0.21} & \textbf{80.88 $\pm$ 0.14} & \textbf{71.22 $\pm$ 0.23} & \textbf{86.03 $\pm$ 0.15} & \textbf{73.40 $\pm$ 0.21} & \textbf{86.51 $\pm$ 0.15} \\
\bottomrule
  \end{tabular}
  }
\end{table*}
\vspace{-10pt}

\subsubsection{Effectiveness of the geodesic distance measure ($d_{g}$)}
The geodesic distance ($d_g$) contributes substantial improvements across all benchmarks. As shown in Table~\ref{tab:fs_ablation}, integrating $d_g$ elevates accuracy by \textbf{3.30\%} (61.07\% $\rightarrow$ 64.37\%), while 5-shot performance gains \textbf{4.70\%} (76.18\% $\rightarrow$ 80.88\%). For tieredImagenet, $d_g$ drives \textbf{3.44\%} (67.78\% $\rightarrow$ 71.22\%) and \textbf{4.10\%} (81.93\% $\rightarrow$ 86.03\%) improvements in 1-shot and 5-shot settings, respectively. These results demonstrate that $d_{g}$ effectively captures hierarchical relationships in hyperbolic space, facilitating accurate class differentiation under limited data conditions. 

% The geodesic distance measure is integral to the HS2SD framework. For few-shot image classification, integrating $d_{g}$ leads to notable accuracy improvements. On miniImagenet, the 1-shot accuracy increases from 61.07\% to 64.37\%, and the 5-shot accuracy increases from 76.18\% to 80.88\%. Similar enhancements are observed on other datasets, such as tieredImagenet (67.78\% to 71.22\% in 1-shot setting and from 81.93\% to 86.03\% in 5-shot setting). These results demonstrate that $d_{g}$ effectively captures hierarchical relationships in hyperbolic space, facilitating accurate class differentiation under limited data conditions. 
% For EM tasks, $d_{g}$ also proves essential. Without $d_{g}$, HS2SD method fails to converge, indicating its critical role in aligning embeddings for accurate matching. When integrated, $d_{g}$ enables the model to achieve best performance across most datasets, further emphasizing its importance in capturing the complex relationships inherent to EM tasks.

\subsubsection{Effectiveness of the topological distance measure ($d_{t}$)}
% \todo{Modify this part}
The incorporation of topological distance ($d_t$) consistently improves the model performance across all datasets, demonstrating its effectiveness in few-shot classification tasks. Notably, $d_t$ exhibits more pronounced improvements in 1-shot scenarios, achieving gains of \textbf{1.37\%} (63.00\% $\rightarrow$ 64.37\%) on miniImagenet and \textbf{1.16\%} (70.06\% $\rightarrow$ 71.22\%) on tieredImagenet, while showing modest gains of 0.39\% and 0.13\% in their respective 5-shot counterparts. This systematic pattern across datasets supports our hypothesis that local feature relationships compensate for prototype uncertainty under extreme data scarcity.

\subsubsection{Effectiveness of $\lambda$ adapter \(\mathcal{G}\)}
The proposed $\lambda$ adapter module (\(\mathcal{G}\)) addresses the inherent challenge of fixed weight assignment in few-shot learning scenarios. As evidenced in Table~\ref{tab:fs_ablation}, our adapter yields consistent improvements across all benchmarks: +1.06\%(1-shot) and +1.22\%(5-shot) on miniImagenet, +1.48\%/+1.91\% on tieredImagenet, and +0.91\%/+1.19\% on CIFAR-FS. This adaptability underscores the robustness of \(\mathcal{G}\) in optimizing \(\lambda\) for diverse scenarios, enhancing the applicability of our method.

\noindent\begin{wrapfigure}{r}{0.3\textwidth} % r=右侧，0.3\textwidth=图片宽度
  \centering
  \includegraphics[width=\linewidth]{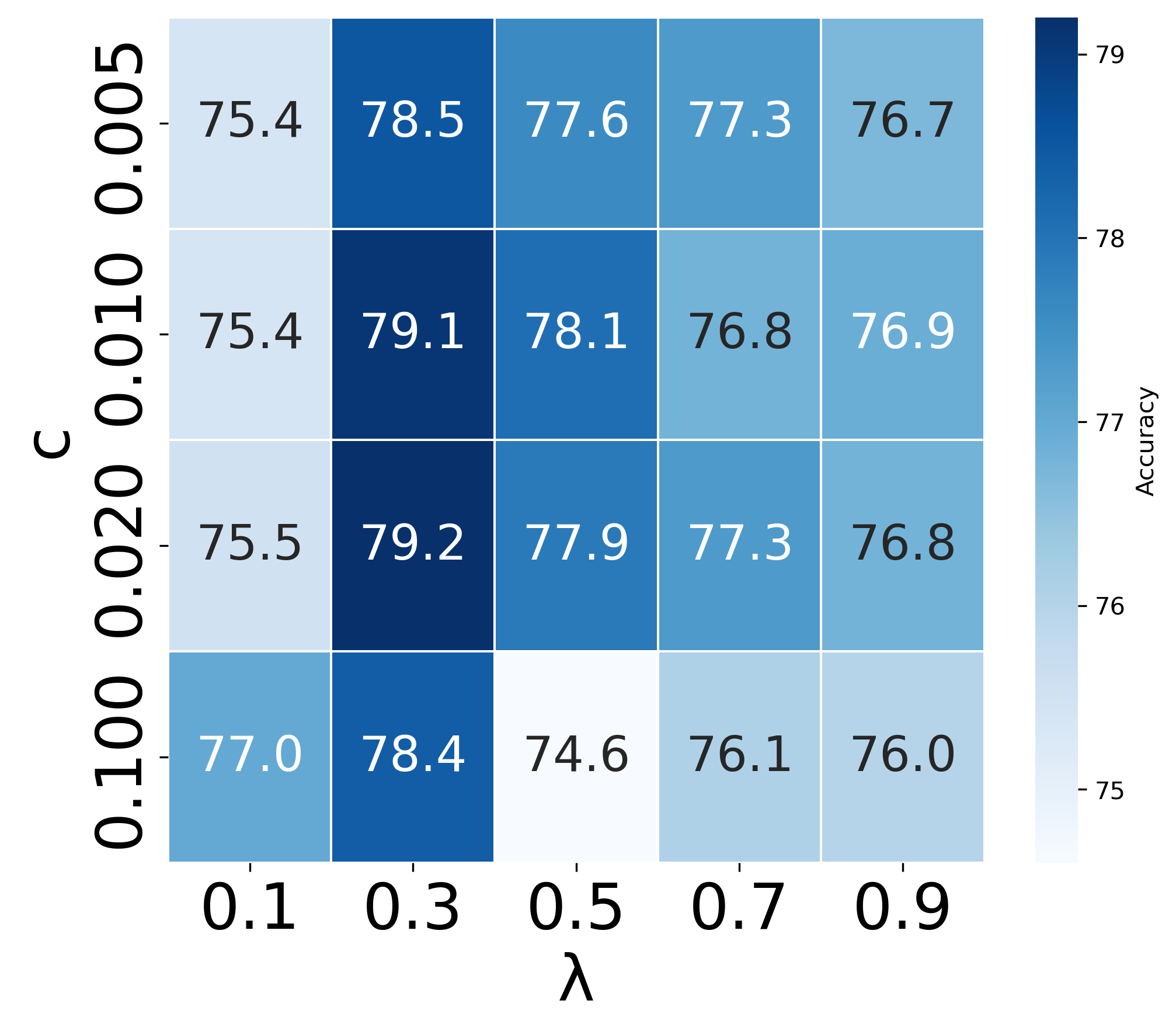} % \linewidth 自动适配 wrapfigure 宽度
  \caption{Ablation study on curvature $c$ and lambda $\lambda$ on AG dataset for EM task.}
  \label{fig:ablation_lambdac}
  \vspace{-30pt}
\end{wrapfigure}

\subsubsection{Ablations on the curvature $c$ and the $\lambda$ of HS2SD}

% To evaluate the impact of curvature \(c\) in the hyperbolic space and the balancing parameter \(\lambda\) in HS2SD, we conduct ablation experiments on AG for EM task and miniImagenet for few-shot image classification task. The results are summarized in Table~\ref{tab:ablation_lambdac} and Table~\ref{tab:ablation_lambdaada}

\noindent\textbf{Impact of Curvature.} Fig.~\ref{fig:ablation_lambdac} highlights the influence of curvature \(c\) on the performance of our method in the EM task. Higher curvature values (e.g., \(c=0.020\)) generally yield better results across different \(\lambda\) values, with the best accuracy of 79.2\% achieved at \(c=0.020\) and \(\lambda=0.3\). This improvement suggests that larger curvatures, which correspond to capturing more intricate hierarchical structures, better align with the dataset's characteristics and the design of our method.

\noindent\textbf{Impact of $\lambda$.}
The parameter \(\lambda\) is essential for balancing the contributions of geodesic distance (\(d_g\)) and topological distance (\(d_t\)) in our metric. As shown in Fig.~\ref{fig:ablation_lambdac}, \(\lambda=0.3\) consistently achieves the best performance for all curvature settings, highlighting its effectiveness in optimizing the trade-off between \(d_g\) and \(d_t\). However, extremely low or high \(\lambda\) values (e.g., \(\lambda=0.1\) or \(\lambda=0.9\)) lead to performance degradation. This likely occurs because overemphasizing one distance metric undermines the complementary relationship between \(d_g\) and \(d_t\), reducing the overall effectiveness of the method. 

\subsection{Efficiency and Memory Analysis}

\begin{wrapfigure}{r}{0.4\textwidth}
\vspace{-20pt}
\begin{minipage}[t]{0.4\textwidth}
\captionsetup{type=table}
\captionof{table}{Time (ms) and GPU memory cost (GB) during training and inference phases on miniImagenet 1-shot tasks. We report the average time and peak GPU usage (per iteration) across 1,000 runs.
% Training configurations: Hyp-ProtoNet (20-way training, 5-way inference) and HS2SD (10-way training, 5-way inference).
}
\label{tab:efficiency_analysis}
\centering
\resizebox{\textwidth}{!}{
\begin{tabular}{l|cc|cc}
\toprule
\multirow{2}{*}{\textbf{Method}} & \multicolumn{2}{c|}{\textbf{Training}} & \multicolumn{2}{c}{\textbf{Inference}} \\
& \textit{Time} & \textit{Mem.} & \textit{Time}  & \textit{Mem.} \\
\midrule
Hyp-ProtoNet\citep{khrulkov2020hyperbolic} & 106.17 & 10.25 & 20.02 & 0.50 \\
HS2SD (w/o $d_{t}$) & 74.56 & 7.80 & 25.56 & 0.66 \\
HS2SD (w/o $d_{g}$) & 92.25 & 9.32 & 29.70 & 0.68 \\
HS2SD & 92.31 & 9.34 & 32.06 & 0.69 \\
\bottomrule
\end{tabular}
}
\end{minipage}
\vspace{-10pt}
\end{wrapfigure}

Table \ref{tab:efficiency_analysis} shows our method's computational efficiency. 
For fair comparison, we evaluate both methods using their respective optimal training configurations: Hyp-ProtoNet with $\mathrm{20 \ way}$ and ours with $10 \ \mathrm{way}$ training. During training, our full model uses \textbf{8.9\% less memory} (9.34 GB vs. 10.25 GB) and is \textbf{13.1\% faster} (92.31 ms vs. 106.17 ms per iteration) than Hyp-ProtoNet. The integration of $d_t$ adds minimal overhead during training, with negligible inference costs (4\% memory increase). These results highlight that HS2SD delivers superior performance while maintaining computational efficiency.
 % While the increase in resource consumption is significant, it remains within acceptable limits given the improved accuracy provided by $d_{t}$.

% \px{474 line in cvpr}
% \px{Experimental configuration. } 
% \todo{i will put this on the appendix}
% \px{time retains two decimal places}

% \begin{table}[H]
%     \centering
%     % \resizebox{0.5\textwidth}{!}{
%     \begin{tabular}{lrr} % Added | between the first and second column
%         \toprule
%          \textbf{Method} & \textbf{Time(ms)} & \textbf{GPU Mem(GB)}\\
%         \midrule
%         Baseline &80 & 5.04\\
%         HS2SD (w/o $d_{t}$) & 83 & 5.04\\
%         HS2SD (w/o $d_{g}$) & 99 & 8.40\\
%              HS2SD     & 100 & 8.40\\        
%         \bottomrule
%     \end{tabular}
%     % }
%     \caption{Inference time and memory cost analysis of HS2SD and baseline.}
%     \label{tab:efficiency_analysis}
%     % \vspace{-5pt}
% \end{table}

\section{Conclusion}\label{sec:conclusion}
In this work, we have presented the Hyperbolic Set-to-Set Distance (HS2SD), a distance measure that effectively captures both global and local structures in hyperbolic spaces. By combining geodesic distance for global relationships and topological distance for structural variations, HS2SD demonstrated superior performance in entity matching and few-shot image classification, particularly on datasets with complex hierarchical structures. Theoretical analysis and experiments highlight the unique strengths of hyperbolic spaces in modeling complex data structures and demonstrate the value of combining global and local perspectives for robust distance metrics.

\textbf{Limitation.} Our method assumes a uniform negative curvature throughout the hyperbolic space. In scenarios where data relationships exhibit varying local curvatures, a single global curvature parameter may not capture all nuances, potentially limiting representational flexibility. We will explore adaptive curvature for the HS2SD in the future.
\medskip

{
\small
\bibliographystyle{plain}
\bibliography{reference_header,references}
}

%%%%%%%%%%%%%%%%%%%%%%%%%%%%%%%%%%%%%%%%%%%%%%%%%%%%%%%%%%%%

% APPENDIX
%%%%%%%%%%%%%%%%%%%%%%%%%%%%%%%%%%%%%%%%%%%%%%%%%%%%%%%%%%%%%%%%%%%%%%%%%%%%%%%
%%%%%%%%%%%%%%%%%%%%%%%%%%%%%%%%%%%%%%%%%%%%%%%%%%%%%%%%%%%%%%%%%%%%%%%%%%%%%%%
% \newpage
\appendix
% \maketitle
\onecolumn

\section{Broader Impact}
\label{app:broader_impact}
The ability to leverage both global structure and fine-grained topology makes HS2SD especially valuable in domains with scarce labeled examples—such as rare disease imaging or biodiversity monitoring—by extracting maximal information from limited samples and potentially accelerating scientific discovery.

While HS2SD’s sensitivity to hierarchical patterns can improve discrimination of subtle classes, it may also encode and amplify existing biases present in training data hierarchies (e.g., socio-demographic groupings in user data). Care must be taken to audit embeddings and distance outputs to prevent unfair treatment or reinforcement of stereotypes.

\section{More experiment details} \label{appendix:exp}
\subsection{Few-shot learning}
\subsubsection{Dataset.}
The miniImagenet dataset contains 100 classes from the ImageNet dataset~\cite{ILSVRC15}, containing 600 images for each class. We split the 100 classes into 64, 16, and 20 classes for training, validation, and testing, respectively. The tieredImagenet dataset has 779165 images from 608 classes, where 351, 97, and 160 classes are used for training, validation, and testing, respectively.  The CUB dataset contains 11,788 images from 200 classes, where 200 classes are divided into 100, 50, and 50 for training, validation, and testing, respectively. CIFAR-FS is a few-shot classification dataset built on CIFAR100~\cite{cifar}. It contains 64, 15, and 20 classes for training, validation, and testing, respectively.

\subsection{Standard Image Classification}
\subsubsection{Dataset.}
MNIST contains 10 classes with 70,000 grayscale images of handwritten digits, each with a resolution of 28 × 28 pixels. CIFAR10 comprises 60,000 natural color images of 10 classes, with each image sized at 32 × 32 pixels. CIFAR100 extends to 100 classes with 60,000 natural color images in total, maintaining the same 32 × 32 pixel resolution.
% \subsubsection{Implementation details.}

% The experiment was conducted using an Intel® Xeon® Gold 6326 2.90GHz CPU, two NVIDIA 4090D GPUs, and 512GB of memory.
% We measure the time cost and GPU memory usage of the model from the completion of data loading to the end of the forward pass. The dataset used in the experiment is miniImagenet, with images resized to 84 × 84 pixels before passed to the model. Each metric is measured 200 times, and the mean value is reported.
% \section{Preliminaries}
% \subsection{Lorentz Model}
% \subsection{Poincaré Model}

\section{Resource requirements} \label{appendix:resources}
All experiments were conducted on NVIDIA RTX 4090 GPUs. The few-shot image classification experiments required approximately 0.2 GPU hours per 10 epochs, while entity matching experiments consumed approximately 1.5 GPU hours per 10 epochs.

\section{Proof of \cref{thm-McKay-main}}
\label{app:proof}

Based on the adjacency and degree matrix of graphs, we will try to construct matrices whose spectra may effectively distinguish trees especially. Our construction resorts to the key result of \cite{McKay} and explicit construction of real irreducible representations of matrix algebras generated by adjacency and degree matrices.

\subsection{General results on cospectral graphs}
We call two $n$ by $n$ matrices $\boldsymbol{A},\boldsymbol{B}\in \boldsymbol{M}_n(\mathbb{R})$ \textit{cospectral} if $\det(\boldsymbol{A}-x\boldsymbol{I}_n)=\det(\boldsymbol{B}-x\boldsymbol{I}_n), \forall x$, i.e. they have the same characteristic polynomial. An equivalent definition is $\rm tr(\boldsymbol{A}^i)=\rm tr(\boldsymbol{B}^i). i=1,\dots, n$ (see \cite{DH}). We are mainly interested in adjacency matrices of simple connected graphs $G$, and how they are determined by the spectra of the degree matrices $\boldsymbol{D}(G)$, adjacency matrices $\boldsymbol{A}(G)$, the Laplacians $\boldsymbol{L}(G)=\boldsymbol{D}(G)-\boldsymbol{A}(G)$, the signless Laplacian $|\boldsymbol{L}|(G)=\boldsymbol{D}(G)+\boldsymbol{A}(G)$, or other similar simple matrices. Note that any two isomorphic graphs have the same adjacency matrix, degree matrix, or Laplacian. So speaking of ``determined by" some matrix, we always mean up to isomorphism.

It is easy to see that the spectra of adjacency matrices determine the number of length-$k$ walks between any two vertices for any positive integer $k$. Bipartite graphs can also be characterized because the spectra of their adjacency matrices are symmetric about zero (see \cite{Biggs}). From the spectra of Laplacian matrices, the number of spanning trees can be determined by Kirchhoff's matrix tree theorem (see \cite{OT}). Actually, there are special families of graphs determined by spectra of these simple matrices, such as cycles, paths, star-like graphs, etc. Especially, regular graphs are determined by the spectra of the adjacency matrices of themselves and their complements together with the Laplacian (see Proposition 3 of \cite{DH}).

However, trees as an important and prevalent family of graphs can not be determined by spectra of those simple matrices. Schwenk \cite{Schwenk} first proved that the ratio of trees determined by the spectra of adjacency matrices tends to 0 as the number of vertices increases. Godsil and McKay \cite{GM} later proved that the ratio tends to zero even for trees determined by the spectra of their adjacency matrices and those of their complement graphs. The results have also been generalized to more matrices by McKay \cite{McKay} as follows.

\begin{theorem}[Corollary 4.4 of \cite{McKay}]
    Let $ p(n) $ be the proportion of trees $ T $ on $n$ vertices which are determined by the spectra of the adjacency matrix $ \boldsymbol{A}(T)$ ($\boldsymbol{A}(\overline{T})$), the degree matrix $\boldsymbol{D}(T)$ ($\boldsymbol{D}(\overline{T})$), the Laplacian $\boldsymbol{L}(T)$ ($\boldsymbol{L}(\overline{T})$) , or the signless Laplacian $ |\boldsymbol{L}|(T)$ ($|\boldsymbol{L}|(\overline{T})$), where $ \overline{T} $ denotes the complement graph of $ T $. Then $ p(n) \rightarrow 0 $, as $ n \rightarrow \infty $.
\end{theorem}
Thus, different from regular graphs, to distinguish trees (up to isomorphism), computing the spectrum of any single commonly used matrix is not enough.

\subsection{The distinction of trees and structure of matrix algebras}
 To address the issue, McKay provided an efficient criterion of determining trees by the adjacency and degree matrix altogether as follows.

\begin{theorem}[Theorem 5.3 of \cite{McKay}]\label{thm-McKay}
	Two trees $ T_1 $ and $ T_2 $ are isomorphic if and only if for any two variable polynomial $p$ (non-commutative), $ p(\boldsymbol{A}(T_1), \boldsymbol{D}(T_1)) $ and $ p(\boldsymbol{A}(T_2), \boldsymbol{D}(T_2)) $ are cospectral.
\end{theorem}

Thus, to check whether two trees are isomorphic, it suffices to compute the spectra of polynomials in the adjacency and degree matrix. However, this results in the problem of efficient computing: is there a finite canonical set of polynomials certified for this purpose? In other words, how can we choose the right polynomials to check? 

To address the issue at least for most trees if not for all, we need to study the structure of matrix sub-algebras of $\boldsymbol{M}_n(\mathbb{R})$ consisting of all polynomials in $\boldsymbol{A}(T)$ and $\boldsymbol{D}(T)$, i.e. of the form: 
\begin{align}\label{eq-def of matrix algebra}
    \mathcal{A}(T)&=\langle \boldsymbol{A}(T), \boldsymbol{D}(T)\rangle\\
    &=\left\{\sum\limits_{w}a_{w}w(\boldsymbol{A}(T),\boldsymbol{D}(T))\mid a_{w}\in\mathbb{R}\right\}\notag
\end{align}
for any tree $T$, where $w(\boldsymbol{A},\boldsymbol{D})$ denotes any binary word in the alphabet $\{\boldsymbol{A}, \boldsymbol{D}\}$. We will see that it is convenient to study such matrix algebras from the perspective of representation theory. To be clear, by a \textit{representation} we always mean a left module of an algebra. 

Clearly, any algebra $\mathcal{A}\subset \boldsymbol{M}_n(\mathbb{R})$ (the $n$ by $n$ matrices on $\mathbb{R}$) is finite dimensional. Then we define its radical ${Rad}(\mathcal{A})$ to be the set of all elements of $\mathcal{A}$ which act by 0 in all irreducible representations of $\mathcal{A}$. Here an irreducible means without proper subrepresentations, i.e. without proper invariant subspaces under the action by $\mathcal{A}$. The radical is actually the maximal nilpotent ideal of $\mathcal A$ (Proposition 2.11 of Etingof et. al. \cite{Etingof}), i.e. ${Rad}(\mathcal A)=\left\{x \in \mathcal A \mid x^n=0 \text { and } \forall y \in \mathcal{A}, (x y)^n=0\right\}$. For matrix algebras generated by symmetric matrices like $\mathcal{A}$, their radical is actually trivial.

\begin{lemma}\label{lem-trivial radical}
For any symmetric matrices $\boldsymbol{A}_1,\dots,\boldsymbol{A}_k\in \boldsymbol{M}_{n\times n}(\mathbb{R})$, $Rad(\langle \boldsymbol{A}_1,\dots, \boldsymbol{A}_k\rangle)=0$.

\end{lemma}
\begin{proof}
$\forall \boldsymbol{B}\in Rad(\langle \boldsymbol{A}_1,\dots, \boldsymbol{A}_k\rangle)$ as a polynomial in $\boldsymbol{A}_i$'s, $\boldsymbol{B}^T$ also belongs to the algebra. Then by Proposition 2.11 of \cite{Etingof}, the symmetric matrix $\boldsymbol{B}\boldsymbol{B}^T$ (or $\boldsymbol{B}^T\boldsymbol{B}$) is nilpotent, i.e. its eigenvalues are all zero. But symmetric nilponent matrix must be zero, so $\boldsymbol{B}\boldsymbol{B}^T=\boldsymbol{B}^T\boldsymbol{B}=0$ which implies $\boldsymbol{B}=0$. More explicitly, every real matrix has singular value decomposition, say $\boldsymbol{B}=\boldsymbol{U}\boldsymbol{\Sigma} \boldsymbol{V}^T$ for some orthogonal matrices $\boldsymbol{U},\boldsymbol{V}$ and diagonal matrix $\boldsymbol{\Sigma} $, so then $\boldsymbol{B}\boldsymbol{B}^T=\boldsymbol{U}\boldsymbol{\Sigma} ^2\boldsymbol{U}^T$. Since the eigenvalues of $\boldsymbol{B}\boldsymbol{B}^T$ are all zero, its similar diagonal matrix $\boldsymbol{\Sigma} ^2=0$, i.e. $\boldsymbol{\Sigma} =0$. Thus $\boldsymbol{B}=0$. 
\end{proof}

Finite dimensional algebras over algebraically closed fields (say $\mathbb{C}$) has the following straightforward structure:

\begin{theorem}[Theorem 2.5 of \cite{Etingof}]\label{thm-repn of algebras}
A finite dimensional algebra $\mathcal A$ has finitely many irreducible representations $W_i$ up to isomorphism, which are finite dimensional, and
$$
\mathcal{A} / {Rad}(\mathcal{A}) \cong \bigoplus_i \mathrm{End}(W_i).
$$
\end{theorem} 
Thus, if we consider $\mathcal{A}$ as matrix algebras over $\mathbb{C}$, together with Lemma \ref{lem-trivial radical} that $Rad(\mathcal{A})=0$, then we simply have
\begin{corollary}\label{cor-semisimple}
    For any connected graph $G$, the algebra $\mathcal{A}(G)=\langle \boldsymbol{A}(G),\boldsymbol{D}(G)\rangle$ is semisimple, i.e. $Rad(\mathcal{A}(G))=0$ and 
    \[\mathcal{A}(G)\cong \bigoplus_i \mathrm{End}(W_i),\]
    summing over all the complex irreducible representations $W_i$.
\end{corollary}

However, noting that McKay's Theorem \ref{thm-McKay} works over $\mathbb{R}$, the above description of matrix algebras $\mathcal{A}$ may not stand even for trees, since some irreducible representation $V_i$ may not be isomorphic to an invariant subspace of $\mathbb{R}^n$ for $\mathcal{A}$. In general, it is not trivial to find irreducible representations of real matrix algebras. We will introduce a natural and explicit construction of special irreducible representations for $\mathcal{A}$.

Without ambiguity, we will always abbreviate $\mathcal{A}(G), \boldsymbol{A}(G), \boldsymbol{D}(G), \boldsymbol{L}(G)$ to $\mathcal{A}, \boldsymbol{A}, \boldsymbol{D}, \boldsymbol{L}$. The Laplacian $\boldsymbol{L}=\boldsymbol{D}-\boldsymbol{A}\in\mathcal{A}$ helps produce the matrix $\boldsymbol{J}$ (with all entries 1). First, simply by

\[x^T(\boldsymbol{\Lambda}-\boldsymbol{A})x=\sum_{(i,j) \text{ is edge}}(x_i-x_j)^2,\ \forall x=(x_1,\dots,x_n)\in\mathbb{R},\] 
we see that the $0$-eigenspace of $\boldsymbol{L}$ is one-dimensional, spanned by $\boldsymbol{1}=(1,\dots,1)\in\mathbb{R}^n$. Thus, $\boldsymbol{L}$ has rank $n-1$. Hence, its minimal polynomial is $P(x)=xQ(x)$, $Q(\boldsymbol{L})=\boldsymbol{1}a^T$ for some $a\in\mathbb{R}^n$ noting that $P(\boldsymbol{L})=\boldsymbol{L}Q(\boldsymbol{L})=\boldsymbol{0}$. Since $\boldsymbol{A}, \boldsymbol{D}$ are symmetric, whence $\boldsymbol{L}$ and $Q(\boldsymbol{L})$, there is
\begin{equation}\label{eq-J}
\boldsymbol{1}a^T=Q(\boldsymbol{L})=Q(\boldsymbol{L})^T=a\boldsymbol{1}^T,
\end{equation}
that is, $a=c\boldsymbol{1}$ $Q(\boldsymbol{L})=c\boldsymbol{1}\boldsymbol{1}^T=c\boldsymbol{J}$ for some $c\in\mathbb{R}$. Thus, $\boldsymbol{J}\in\mathcal{A}$.

The above construction recalls a more general important result as follows.

\begin{theorem}[Theorem 7.3 of Chan-Godsil \cite{CG}]\label{thm-CG}
	Let $G$ be a connected graph and $v_1, \ldots, v_m$ be characteristic vectors of the subsets $S_1, \ldots, S_m$ of $V(G)$ (the vertices of $G$). Let $W$ be the space spanned by all vectors of the form $\boldsymbol{A}^r v_i$. Then the algebra generated by $\boldsymbol{A}$ and $v_iv_i^T(i=1, \ldots, m)$ is isomorphic to the direct sum of $\mathrm{End(W)}$ and the algebra generated by the restriction of $\mathcal{A}$ to $W^{\perp}$, the space orthogonal to $W$.
\end{theorem}
By Lemma \ref{lem-trivial radical}, the algebra generated by $\boldsymbol{A}$ and $v_iv_i^T$ is semisimple, hence $V=\langle \boldsymbol{A}^rv_i\rangle$ is its irreducible representation. 
The proof of Theorem \ref{thm-CG} is based on ideas of Laffey \cite{L}, by which we will prove a similar result for $\mathcal{A}=\langle \boldsymbol{A},\boldsymbol{D}\rangle$. Here we only mention that $v_jv_j^T(\boldsymbol{A}^rv_i)=(v_j^T\boldsymbol{A}^rv_i)v_j$ is a multiple of $v_j$ hence belongs to $V$, which shows that $V$ is truly a representation of $\langle \boldsymbol{A},v_iv_i^T\rangle$. In our case, to make $v_iv_i^T\in\mathcal{A}$, the subsets $S_i$ should be invariant under the automorphism group of the graph $G$, denoted by $\mathrm{Aut}(G)$. We have the following easy representation of $\mathrm{Aut}(G)$:

\begin{lemma}\label{lem-automorphism}
	For any graph $G$ and its adjacency matrix $\boldsymbol{A}$, $\mathrm{Aut}(G)$ consists of permutation matrices that commute with $\boldsymbol{A}$.
\end{lemma}

Clearly degrees of vertices are invariant under automorphisms, hence $\mathrm{Aut}(G)$ also commute with $\boldsymbol{D}$. This immediately shows that 

\begin{corollary}\label{cor-commute}
For any graph $G$, $\langle \boldsymbol{A},\boldsymbol{D}\rangle\subset \boldsymbol{M}_n(\mathbb{R})^{\mathrm{Aut(G)}}$, the subalgebra consisting of matrices fixed by the conjugation action of $\mathrm{Aut(G)}$ on $\boldsymbol{M}_n(\mathbb{R})$. 
\end{corollary}

Due to Corollary \ref{cor-commute}, natural choices of the subsets $S_i\subset V(G)$ as in Theorem \ref{thm-CG} are sets invariant under permutations of $\mathrm{Aut(G)}$. The most obvious ones are just the orbits of $\mathrm{Aut(G)}$ in $V(G)$. 

\begin{definition}\label{def-inv.}
For any simple connected graph $G$, let $\mathcal{O}=\{o_1,\dots,o_M\}$ be the set of orbits of $\mathrm{Aut(G)}$ in vertices of $G$. For each $o_j\in \mathcal{O}$, let $u_j=\frac{1}{\sqrt{|o_j|}}\boldsymbol{1}_{o_j}$ be the normalized characteristic vector (which may also be called orbit) of $o_j$ and $\boldsymbol{J}_j=u_ju_j^T$. Denote $\mathcal{A}_{O}=\langle \boldsymbol{A}, \boldsymbol{J}_j\rangle_{o_j\in\mathcal{O}}$, the algebra generated by $\boldsymbol{A}$ and $\boldsymbol{J}_j$'s.
\end{definition}

\begin{lemma}\label{lem-orbits}
 The vector space $U=\langle u_j\rangle$ is a representation of $\mathcal{A}$ and $\mathcal{A}_O$.
\end{lemma}
\begin{proof}
    Note that the permutations of $\mathrm{Aut}(G)$ commute with $\boldsymbol{A}$ and $\boldsymbol{D}$ and $\forall \boldsymbol{P}\in\mathrm{Aut}(G)$, $\boldsymbol{P}u_j=u_j$. Hence
    $$\boldsymbol{P}\boldsymbol{D} u_j=\Lambda \boldsymbol{P}u_j=\boldsymbol{D} u_j, \boldsymbol{P}\boldsymbol{A}u_j=\boldsymbol{A}\boldsymbol{P}u_j=\boldsymbol{A}u_j,$$
    i.e. $\boldsymbol{D}u_j$ and $\boldsymbol{A}u_j$ are invariant under the action of $\rm{Aut}(G)$, which shows that they are all linear combinations of orbits. Thus $U=\langle u_j\rangle$ is an invariant subspace, i.e. a representation of $\mathcal{A}$. Also for any $k$, $\boldsymbol{J}_ku_j=u_ku_k^Tu_j=(u_k^Tu_j)u_k\in U$. Hence $U$ is also a representation of $\mathcal{A}_O$.
\end{proof}

To further study the relation of $\mathcal{A}$ and $\mathcal{A}_O$, we resort to Theorem \ref{thm-CG} to prove
\begin{lemma}\label{lem-J's}
	Provided the notations above, $$\ \mathcal{A}_O=\mathrm{End}(U)\oplus\langle \boldsymbol{A}_{U^{\perp}}\rangle,$$
    where $U=\langle u_j\rangle=\langle \boldsymbol{A}^ru_j\rangle_{r\in\mathbb{Z}_{\geq0}, o_j\in\mathcal{O}}$ is an irreducible representation of $\mathcal{A}_O$, and  $\boldsymbol{A}_{U^{\perp}}$ is the restriction of $\boldsymbol{A}$ to the orthogonal space of $U$.
\end{lemma}
\begin{proof}
	The direct sum decomposition is due to Theorem \ref{thm-CG}, for $U=\langle u_j\rangle=\langle \boldsymbol{A}^ru_j\rangle_{r\in\mathbb{Z}_{\geq0}, o_j\in\mathcal{O}}$ by Lemma \ref{lem-orbits}. By Lemma \ref{lem-trivial radical}, $\mathcal{A}_O$ is semisimple and so $U$ is irreducible.
\end{proof}

Now, resorting to the ideas of \cite{L} used in proving Theorem \ref{thm-CG} (see \cite{CG}), we now prove the following

\begin{proposition}\label{prop-1st irreducible repn}
    $U=\langle u_j\rangle=\mathcal{A}\boldsymbol{1}_n$ is an irreducible representation of $\mathcal{A}=\mathcal{A}(T)$ for any tree $T$.
\end{proposition}
As we shall show, $\mathcal{A}$ is irreducible for any graphs. However, it might only be a subrepresentation of $U$ ``supported" on an equitable partition in general. To prove the proposition above, we need the following key lemmas. First, we introduce the promised explicit construction of special irreducible representations for $\mathcal{A}$.

\begin{lemma}\label{lem-rank 1}
For any graph $G$ on $n$ vertices and $\mathcal{A}=\mathcal{A}(G)$, if $a\in\mathcal{A}$ is of rank 1, which must be of the form $a=xy^T$ for some $x,y\in\mathbb{R}^n$, then $x,y\in U$ and $\mathcal{A}x=\mathcal{A}y\subset U$ is an irreducible representation of $\mathcal{A}$. In particular, $\mathcal{A}\boldsymbol{1}$ is an irreducible representation of $\mathcal{A}$.
\end{lemma}

\begin{proof}
    If $xy^T\in\mathcal{A}$, then $(xy^T)^T=yx^T\in\mathcal{A}$ since $\mathcal{A}$ is symmetric. Thus, clearly $\mathcal{A}x\subset\mathcal{A}y$ and $\mathcal{A}y\subset\mathcal{A}x$, i.e., $\mathcal{A}x=\mathcal{A}y$. Also, $xy^Tyx^T=\|y\|^2xx^T\in\mathcal{A}$ and $yx^Txy^T=\|x\|^2yy^T\in\mathcal{A}$. Moreover, for any automorphism $\boldsymbol{P}\in\mathrm{Aut}(G)$, $\boldsymbol{P}xx^T=xx^T\boldsymbol{P}=x(\boldsymbol{P}^Tx)^T$, which shows that $\boldsymbol{P}x=x$. Hence, $x\in U$ and $\mathcal{A}x\subset U$. 
    In addition, we simply have $\mathrm{End}(\mathcal{A}x)=\mathcal{A}xx^T\mathcal{A}\subset\mathcal{A}$. Hence, $\mathcal{A}x$ is an irreducible representation of $\mathcal{A}$.

    In particular, we know that $\boldsymbol{J}=\boldsymbol{1}\boldsymbol{1}^T\in\mathcal{A}$ by (\ref{eq-J}). Thus, for any $a,b\in\mathcal{A}$, $(a\boldsymbol{1})(b\boldsymbol{1})^T=a\boldsymbol{J}b^T\in\mathcal{A}$, noticing that $b^T=p(\boldsymbol{A},\boldsymbol{D})^T\in\mathcal{A}$ if $b=p(\boldsymbol{A},\boldsymbol{D})$ for some polynomial $p$. Thus, $\mathrm{End}(\mathcal{A}\boldsymbol{1})\subset\mathcal{A}$, so that $\mathcal{A}\boldsymbol{1}$ must be irreducible.
\end{proof}
Especially, for any simple eigenvector $v$ of any matrix in $\mathcal{A}$, $\mathcal{A}v\subset U$ is an irreducible representation of $\mathcal{A}$.

\begin{lemma}[Lemma 5.1 of \cite{McKay}]\label{lem-distance matrix}
    For any tree $T$ on $n$ vertices, the distance matrices $(\boldsymbol{\Delta}_r)_{n\times n}\in\mathcal{A}(T)$ for any nonnegative integer $r$, whose $(i,j)$-th entry is $1$ if the distance between vertex $i$ and $j$ is $r$, otherwise 0. 
\end{lemma}
\begin{proof}
    $\boldsymbol{\Delta}_0=\boldsymbol{I}_n, \boldsymbol{\Delta}_1=\boldsymbol{A}, \boldsymbol{\Delta}_2=\boldsymbol{A}^2-\boldsymbol{D}$, and for $r\geq 2$
    \[\boldsymbol{\Delta}_{r+1}=\boldsymbol{A}\boldsymbol{\Delta}_{r}-(\boldsymbol{A}-\boldsymbol{I}_n)\boldsymbol{\Delta}_{r-1},\]
    which are clearly all polynomials in $\boldsymbol{A}$ and $\boldsymbol{D}$.
\end{proof}

Thus, we always have $\mathcal{A}=\langle\boldsymbol{\Delta}_i\rangle$. Moreover, we need the following basic fact about the center of trees.

\begin{lemma}[\cite{Jordan}]\label{lem-center}
    Let the center $C(T)$ of a tree $T$ be the set of vertices with the smallest eccentricity, that is, 
    \[C(T)=\mathrm{arg}\min_{v\in T}\max_{u\in T}d(v,u).\]
    Then $C(T)$ consists of one vertex or two vertices joined by an edge, which are symmetric with respect to any automorphism.
\end{lemma}

$\mathbf{Claim}$:
    For any tree $T$ and $\mathcal{A}=\mathcal{A}(T)$, $\mathcal{A}\boldsymbol{1}$ identifies the center $C(T)$ of $T$, that is, $\boldsymbol{1}_{C(T)}\in\mathcal{A}\boldsymbol{1}$. 
\iffalse
\begin{proof}
    Specifically, the leaf set $V_1$ serves as the bottom (first) layer with $\boldsymbol{1}_{V_1}\in\mathcal{A}\boldsymbol{1}$, the second layer $V_2$ is the leaf set of the tree $T_1$ with $V_1$ trimmed off from $T$ (i.e., the support of $A\boldsymbol{1}_{V_1}$) and $\boldsymbol{1}_{V_2}\in\mathcal{A}\boldsymbol{1}$; in sequence, the $k$-th layer $V_k$ is the leaf set of $T_{k-1}$ with $V_{k-1}$ trimmed off from $T_{k-1}$ and $\boldsymbol{1}_{V_k}\in\mathcal{A}\boldsymbol{1}$ for $k\geq 2$; the top layer is the center of $T$. Let $r=\max_{u\in T}d(u,C(T))$ be the eccentricity of the center of $T$. Then clearly $r$ is the distance between the leaves $V_1$ of $T$ and $C(T)$. Thus, the union of the supports of $\boldsymbol{\Delta}_i\boldsymbol{1}_{V_1}$ grows with $i$ and covers $V(T)$ until $i=r$ when $C(T)$ is included in the support of $\boldsymbol{\Delta}_r\boldsymbol{1}_{V_1}$. Hence, the vectors $\boldsymbol{\Delta}_i\boldsymbol{1}_{V_1}$ are linearly independent, but not on the restriction outside of $C(T)$. This implies that there is a linear combination $\boldsymbol{\Delta}_r\boldsymbol{1}_{V_1}+\sum_{i=0}^{r-1}a_i\boldsymbol{\Delta}_i\boldsymbol{1}_{V_1}=c\boldsymbol{1}_{C(T)}$ for some positive integer $c$. Thus, $\boldsymbol{1}_{C(T)}\in\mathcal{A}\boldsymbol{1}$.
\end{proof}
\fi

We will not prove the claim here. The claim together with Lemma \ref{lem-rank 1} implies the following. 

\begin{corollary}\label{cor-Ax=A1}
    For any tree $T$, let $\mathcal{A}=\mathcal{A}(T)$, $C(T)$ be the center of $T$ and $xy^T\in\mathcal{A}$. Then $\mathcal{A}x=\mathcal{A}x=\mathcal{A}\boldsymbol{1}=\mathcal{A}\boldsymbol{1}_{C(T)}$ is the unique real irreducible representation of $\mathcal{A}$.
\end{corollary}
\begin{proof}
    By the claim above, $\boldsymbol{1}_{C(T)}\in\mathcal{A}$ so that $\boldsymbol{1}_{C(T)}\boldsymbol{1}_{C(T)}^T\in\mathrm{End}(\mathcal{A}\boldsymbol{1}
    )\subset\mathcal{A}$ and $\mathcal{A}\boldsymbol{1}_{C(T)}=\mathcal{A}\boldsymbol{1}$ due to the irreducibility of the latter. 
    
    For any $x\in U$, since $T$ is connected, there exists $\boldsymbol{B}\in\mathcal{A}$ such that the support of $Bx\in U$ contains $C(T)$. Then $\boldsymbol{1}_{C(T)}^TBx\neq 0$ so that $\boldsymbol{1}_{C(T)}\in\mathcal{A}x$ and $\mathcal{A}\boldsymbol{1}_{C(T)}\subset\mathcal{A}x$ due to the irreducibility of the former. If $xt^T\in\mathcal{A}$, by Lemma \ref{lem-rank 1}, $\mathcal{A}x$ is also irreducible so that $\mathcal{A}\boldsymbol{1}_{C(T)}=\mathcal{A}x$. Furthermore, by Corollary \ref{cor-semisimple}, we see that $\mathcal{A}\boldsymbol{1}=\mathcal{A}\boldsymbol{1}_{C(T)}$ is the only (isomorphism class of) real irreducible representation of $\mathcal{A}$.
\end{proof}

Now we are ready to prove Proposition \ref{prop-1st irreducible repn}.

\begin{proof}[Proof of Proposition \ref{prop-1st irreducible repn}]
    \iffalse
    First, since any invariant space of $\mathcal{A}$ is a direct sum of (sub-)eigenspaces of $\boldsymbol{A}$, any irreducible sub-representation $W\leq\mathcal{A}\boldsymbol{1}_n$ contains an eigenvector $x$ of $\boldsymbol{A}$. If $x$ were orthogonal to every $u_j, \forall o_j\in\mathcal{O}$, then $x$ would be orthogonal to the whole space $U\supset\mathcal{A}\boldsymbol{1}_n$. Thus $u_j^Tx\neq0$ for some $j$ and $u_j\in W$ by irreducibility of $W$ and semisimplicity of $\mathcal{A}$.

    But since $G$ is connected, for any $o_{j'}\in\mathcal{O}(G)$, there exists $r$ such that $u_{j'}^T\boldsymbol{A}^ru_j>0$, which indicates $u_{j'}\in W$. Thus $\mathcal{A}\boldsymbol{1}_n\supset W\supset U$ so that $U=\mathcal{A}\boldsymbol{1}_n$ must be irreducible. \fi

    Suppose $U$ is not irreducible so that $U=\mathcal{A}\boldsymbol{1}_{C(T)}\oplus W$ for some subrepresentation of $W\leq U$ of $\mathcal{A}$, which is a direct sum of indecomposible representations of $\mathcal{A}$. Then there exists $w\in W$ whose support contains $C(T)$, whence $\boldsymbol{1}_{C(T)}^Tw\neq 0$. Thus, $\mathcal{A}\boldsymbol{1}_{C(T)}\leq \mathcal{A}w$ as a subrepresentation. However, this contradicts with $U=\mathcal{A}\boldsymbol{1}_{C(T)}\oplus W$. 
\end{proof}

Although mentioned in the proof of Corollary \ref{cor-Ax=A1}, we summarize the above results on the structure of $\mathcal{A}$ as follows.

\begin{proposition}\label{prop-structure of A}
    For any tree $T$ and $\mathcal{A}=\mathcal{A}(T)$, 
    \[\mathcal{A}=\mathrm{End}(U)\bigoplus_{i}\mathrm{End}(W_i)=(\mathcal{A}\boldsymbol{J}\mathcal{A})\bigoplus_{i}\mathrm{End}(W_i),\]
    where $U=\mathcal{A}\boldsymbol{1}$ and $W_i$ are all irreducible complex representations of $\mathcal{A}$. Especially, all rank one matrices of $\mathcal{A}$ lie in $\mathrm{End}(U)=\mathcal{A}\boldsymbol{J}\mathcal{A}$.
\end{proposition}

\begin{proof}
   By Corollary \ref{cor-semisimple}, $\mathcal{A}$ is the direct sum of endomorphism rings of irreducible real and complex representations. By Lemma \ref{lem-rank 1}, all real irreducible representations are subrepresentations of $U$ while $U=\mathcal{A}\boldsymbol{1}$ itself is irreducible by Proposition \ref{prop-1st irreducible repn}. Hence, $\mathrm{End}(U)=\mathcal{A}\boldsymbol{J}\mathcal{A}$ and the other irreducible reprsentations are all complex. 
\end{proof}

In particular, if $T$ is asymmetric, we have the following.
\begin{corollary}\label{cor-asymmetric}
    For any asymmetric tree $T$, its graph algebra $\mathcal{A}(T)=\langle \boldsymbol{A},\boldsymbol{D}\rangle=\boldsymbol{M}_n(\mathbb{R})$.
\end{corollary}
\begin{proof}
    If $T$ is asymmetric, $U=\mathbb{R}^n$ and $\mathrm{End}(U)=\mathcal{A}\boldsymbol{J}\mathcal{A}=\boldsymbol{M}_n(\mathbb{R})$. Thus, $\mathcal{A}$ must be the entire matrix ring and has no complex irreducible representations.
\end{proof}

This henceforth helps clarify the relation between $\mathcal{A}_O$ and $\mathcal{A}$ as follows.
\begin{corollary}\label{cor-A_O}
As in Definition \ref{def-inv.}, $\mathcal{A}_O\leq\mathcal{A}$ is a subalgebra.
\end{corollary}
\begin{proof}
    By Proposition \ref{prop-1st irreducible repn}, $\forall o_j\in\mathcal{O}, u_j=\boldsymbol{M}_j\boldsymbol{1}_n$ for some $\boldsymbol{M}_j\in\mathcal{A}$. Thus $$\boldsymbol{J}_j=u_ju_j^T=\boldsymbol{M}_j\boldsymbol{1}\boldsymbol{1}^T\boldsymbol{M}_j^T=\boldsymbol{M}_j\boldsymbol{J}\boldsymbol{M}_j^T\in\mathcal{A},$$
    noting that $\boldsymbol{M}_j^T\in\mathcal{A}$ by the symmetry of $\mathcal{A}$.
\end{proof}

\iffalse
In addition, as a representation of $\mathcal{A}_O$, $U^\perp$ is easily pictured as follows. 

\begin{lemma}\label{lem-U' on O}
    $U^\perp$ is a direct sum of 1-dimensional representations of $\mathcal{A}_O$ spanned by eigenvectors of $\boldsymbol{A}$ that are orthogonal to $u_j,\forall o_j\in\mathcal{O}$. 
\end{lemma}
\begin{proof}
    Now that $\boldsymbol{J}_j$ acts on $U^\perp$ as $\boldsymbol{0}$ and by symmetry the invariant spaces of $\mathcal{A}_O$ are direct sums of (sub-)eigenspaces of $\boldsymbol{A}$, the lemma follows immediately.
\end{proof}
This indicates that, under comparatively mild condition, $\mathcal{A}_O=\mathcal{A}$ for almost all graphs, since, roughly speaking, the condition that $\boldsymbol{A}$ has eigenvectors orthogonal to all $u_j$ cuts off a lower dimensional subvariety of the moduli space of all $\mathcal{A}$. We summarize it for trees without rigorous proof as follows.

\begin{corollary}\label{cor-A_O=A}
    For almost all trees, $\mathcal{A}_O=\mathcal{A}$.
\end{corollary}
\fi

We will focus on the algebra $\mathcal{A}_O$ in the next section, where we prove McKay's theorem (Theorem \ref{thm-McKay}).

\subsection{A sketched establishment of McKay's theorem}
Let $T_1, T_2$ be two trees on $n$ vertices such that for any two variable polynomial $p$, $p(\boldsymbol{A}(T_1), \boldsymbol{D}(T_1))$ and $p(\boldsymbol{A}(T_2), \boldsymbol{D}(T_2))$ have the same spectrum. We will show that to identify the orbits of $T_i$, which largely determine whether the trees are isomorphic, we may only focus on the algebra $\mathcal{A}_O=\langle \boldsymbol{A}(T_i), \boldsymbol{J}_j\rangle$, where $\boldsymbol{J}_j=\boldsymbol{1}_{u_j}\boldsymbol{1}_{u_j}^T$ and $u_j$ are orbits of $T_i$ under $\mathrm{Aut}(T_i)$.

Suppose $u_j=p_j(\boldsymbol{A}_1, \boldsymbol{D}_1)\boldsymbol{1}$ for some polynomial $p_j$, then $$\boldsymbol{J}_j=p_j(\boldsymbol{A}_1, \boldsymbol{D}_1)\boldsymbol{J}p_j(\boldsymbol{A}_1, \boldsymbol{D}_1)^T.$$ Noting that by (\ref{eq-J}) $\boldsymbol{J}=Q(\boldsymbol{L}_1)=Q(\boldsymbol{D}_1-\boldsymbol{A}_1)$ is a polynomial in the Laplacian $\boldsymbol{L}_1=\boldsymbol{L}(T_1)$, the polynomials $Q$ and $p_j, p_j^T$ determine $\mathcal{A}_O(T_1)$. Here $p_j^T$ denote the polynomials such that $p_j^T(\boldsymbol{A},\boldsymbol{D})=p_j(\boldsymbol{A},\boldsymbol{D})^T$, provided that transposing a monomial in $\boldsymbol{A}$ and $\boldsymbol{D}$ is just reversing their order by symmetry. Also, noting that $\frac{1}{n}\boldsymbol{J}$ is and idempotent since $\frac{1}{n^2}\boldsymbol{J}^2=\frac{1}{n}\boldsymbol{J}$, its spectrum is $\{1,0,\dots,0\}$. Then by assumption, $\frac{1}{n}Q(\boldsymbol{L}_2)$ must be also an idempotent with the same spectrum. Clearly, this is also the case for all $\frac{1}{|u_j|}\boldsymbol{J}_j$. Although since $\boldsymbol{L}_1$ is similar to $\boldsymbol{L}_2$, we simply have $J=Q(\boldsymbol{L}_1)=Q(\boldsymbol{L}_2)$ by (\ref{eq-J}), the observation on idempotent will be key to our argument.

Let $U_1=\langle u_j\rangle$ and  $U_2=\langle u'_j\rangle$ be the unique real irreducible representation of $\mathcal{A}_i=\mathcal{A}_i(T_i)$ respectively for $i=1,2$, as in Definition \ref{def-inv.}. By Proposition \ref{prop-1st irreducible repn}, we know $U_i=\mathcal{A}_i\boldsymbol{1}$. Now if $p_j(\boldsymbol{A}_1,\boldsymbol{D}_1)\boldsymbol{1}=u_j$, what about $p_j(\boldsymbol{A}_2,\boldsymbol{D}_2)\boldsymbol{1}$? To answer this key question, we first establish the following.

\begin{lemma}\label{lem-dim}
    With the notations above and the same condition with Theorem \ref{thm-McKay}, $\dim(U_1)=\dim(U_2)$. That is, the number of orbits is the same for $T_1$ and $T_2$ under automorphisms. 
\end{lemma}

\begin{proof}
    By abusing notions, we use $u_j$ for the characteristic vector of the orbits or themselves. Since $u_ju_j^T=p_j(\boldsymbol{A}_1,\boldsymbol{D}_1)Q(\boldsymbol{L}_1)p_j(\boldsymbol{A}_1,\boldsymbol{D}_1)^T$ is a scalar of idempotent with spectrum $\{|u_j|,0,\dots,0\}$, by condition we know $p_j(\boldsymbol{A}_2,\boldsymbol{D}_2)Q(\boldsymbol{L}_2)p_j(\boldsymbol{A}_2,\boldsymbol{D}_2)^T=w_jw_j^T$ with $w_j=p_j(\boldsymbol{A}_2,\boldsymbol{D}_2)\boldsymbol{1}$ is similar. Moreover, by $u_ju_j^Tu_{j'}u_{j'}^T=\boldsymbol{0}$ for any $j'\neq j$, we know that $w_jw_j^Tw_{j'}w_{j'}^T=\boldsymbol{0}$. Thus, $w_j\in U_2$ are orthogonal to each other so that $\dim(U_1)\geq\dim(U_2)$. On the contrary, we also have $\dim(U_1)\leq\dim(U_2)$ by a similar argument. Hence, $\dim(U_1)=\dim(U_2)$.
\end{proof}

In the above proof, we see that $\|w_j\|=|u_j|$ by comparing the spectra. For any two variable polynomial $R$, suppose $R(\boldsymbol{A}_1,\boldsymbol{D}_1)u_j=a_{jk}u_k$ and $R(\boldsymbol{A}_2,\boldsymbol{D}_2)w_j=b_{jk}w_k$. Then $u_k^TR(\boldsymbol{A}_1,\boldsymbol{D}_1)u_j=a_{jk}$ and $w_k^TR(\boldsymbol{A}_2,\boldsymbol{D}_2)w_j=b_{jk}$ so that $a_{jk}=b_{jk}$ by similarity of the spectra. Thus, the above arguments actually prove the following.

\begin{corollary}\label{cor-isom}
    Provided the notations above, the two representations $U_1$ and $U_2$ are isomorphic through $u_j\mapsto w_j$ and $\mathrm{End}(U_1)=\mathcal{A}_1\boldsymbol{J}\mathcal{A}_1\simeq\mathcal{A}_2\boldsymbol{J}\mathcal{A}_2=\mathrm{End}(U_2)$.
\end{corollary}

Moreover, we can prove the following.

\begin{lemma}\label{lem-equal sizes of orbits}
    Provided the notations above and conditions of Theorem \ref{thm-McKay} for two trees $T_1$ and $T_2$, the orbits $u_j$ and $\mu_j$ of $T_1$ and $T_2$ respectively are identical under some permutation fixing degrees of the trees.
\end{lemma}

\begin{proof}
    Following the proof of Lemma \ref{lem-dim}, $\frac{1}{\|w_j\|}w_j$ form an orthonormal basis of $U_2$. Thus $\sum_{j}\frac{1}{\|w_j\|}w_jw_j^T$ is the unique idempotent projecting $\mathbb{R}^n$ onto $U_2$, i.e.,
    $$\sum_{j}\frac{1}{\|w_j\|}w_{j}w_{j}^T=\sum_{j}\frac{1}{|\mu_j|}\mu_j\mu_j^T.$$
    Then by similarity of spectra, we see
    $$\sum_{j}\frac{1}{|u_j|}u_ju_j^T\sim\sum_{j}\frac{1}{|\mu_j|}\mu_j\mu_j^T.$$
    Then further by similarity of spectra, this shows  $\sum_{j}u_ju_j^T\sim\sum_{j}\mu_j\mu_j^T$ so that $u_j=\mu_j$ under some permutation. Moreover, by multiplying $\boldsymbol{D}_k(T_i)$ we also know that there is the same number of orbits of any fixed degree in $T_1$ and $T_2$.
\end{proof}

Now we can align the orbits of trees as follows.

\begin{proposition}\label{prop-aligning orbits}
    Provided the notations above and conditions of Theorem \ref{thm-McKay} for two trees $T_1$ and $T_2$, there exist polynomials $p_j$ to identify orbits (under some permutation)of $T_1$ and $T_2$ at the same time.
\end{proposition}
\begin{proof}
    As in the proof of Lemma \ref{lem-dim} and \ref{lem-equal sizes of orbits}, we have for any $x_j\in\mathbb{R}$
    \begin{align*}&\sum_jx_jp_j(\boldsymbol{A}_1, \boldsymbol{D}_1)\boldsymbol{J}p_j(\boldsymbol{A}_1,\boldsymbol{D}_1)^T=\sum_{j}x_ju_ju_j^T\\
    =&\sum_jx_jp_j(\boldsymbol{A}_2, \boldsymbol{D}_2)\boldsymbol{J}p_j(\boldsymbol{A}_2,\boldsymbol{D}_2)^T=\sum_jx_j\mu_j\mu_j^T,\end{align*}
    in which we reorder indices so that $u_j=\mu_j$. Thus, 
    \[p_j(\boldsymbol{A}_1,\boldsymbol{D}_1)\boldsymbol{1}=u_j, p_j(\boldsymbol{A}_2, \boldsymbol{D}_2)\boldsymbol{1}=\mu_j.\]
\end{proof}

\begin{remark}
    Note that we can identify centers by employing distance matrices. As in the proof of Lemma \ref{lem-center}, we let the distance matrices of $T_1$ be $\boldsymbol{\Delta}_{i+1}=\boldsymbol{A}_1\boldsymbol{\Delta}_{i}-(\boldsymbol{A}_1-\boldsymbol{I})\boldsymbol{\Delta}_{i-1}$ and those of $T_2$ be $\Delta'_i$, respectively. Note that they share the same polynomials in $\boldsymbol{A}_i,\boldsymbol{D}_i$. Then by the same linear cancelation argument and similarity of the idempotent spectra, we see that $$\boldsymbol{1}_{C(T_1)}\boldsymbol{1}_{C(T_1)}^T\sim R(\boldsymbol{A}_i,\boldsymbol{D}_i)\boldsymbol{J}R(\boldsymbol{A}_i,\boldsymbol{D}_i)^T\sim \boldsymbol{1}_{C(T_2)}\boldsymbol{1}_{C(T_2)}^T,$$
for some polynomial $R$ and $|C(T_1)|=|C(T_2)|$. Thus, we must have $p\boldsymbol{1}_{C(T_1)}=\boldsymbol{1}_{C(T_2)}$ for some permutation $p$.
\end{remark} 

Proposition \ref{prop-aligning orbits} helps improve Corollary \ref{cor-isom} as follows.

\begin{corollary}\label{cor-align}
    Provided the notations above, the two representations $U_1=U_2$ through $u_j=p\mu_j$ and $\mathrm{End}(U_1)=\mathcal{A}_1\boldsymbol{J}\mathcal{A}_1=p\mathcal{A}_2\boldsymbol{J}\mathcal{A}_2p^T=p\mathrm{End}(U_2)p^T$ for some permutation $p$.
\end{corollary}

In particular, there is a polynomial $R$ such that $R(\boldsymbol{A}_1,\boldsymbol{D}_1)=\boldsymbol{I}_{1}$ and $pR(\boldsymbol{A}_2,\boldsymbol{D}_2)p^T=\boldsymbol{I}_{2}$, where $\boldsymbol{I}_{i}=\boldsymbol{I}_{U_i}$ is the idempotent projecting $\mathbb{R}^n$ onto $U_i$. Since $\boldsymbol{A}_i|_{U_i^\perp}=(\boldsymbol{I}-\boldsymbol{I}_i)\boldsymbol{A}_i$, which is in the expression of a common polynomial ($x-Rx$), $\boldsymbol{A}_i|_{U_i^\perp}$ are similar. Moreover, since $\boldsymbol{I}_i\boldsymbol{A}_i\in\mathrm{End}(U_i)$, we know that 
\begin{align}
    &p\boldsymbol{A}_2|_{U_2^\perp}p^T=p(\boldsymbol{I}-\boldsymbol{I}_2)\boldsymbol{A}_2p^T\notag\\
    =&p\boldsymbol{A}_2p^T-p\boldsymbol{I}_2\boldsymbol{A}_2p^T=p\boldsymbol{A}_2p^T-\boldsymbol{I}_1\boldsymbol{A}_1\notag\\
    =&p\boldsymbol{A}_2p^T-\boldsymbol{A}_1+\boldsymbol{A}_1-\boldsymbol{I}_1\boldsymbol{A}_1\notag\\
    =&(p\boldsymbol{A}_2p^T-\boldsymbol{A}_1)+\boldsymbol{A}_1|_{U_1^\perp},\notag
\end{align}
that is,
\begin{equation}\label{eq-perp}
    p\boldsymbol{A}_2p^T-\boldsymbol{A}_1=p\boldsymbol{A}_2|_{U_2^\perp}p^T-\boldsymbol{A}_1|_{U_1^\perp}.
\end{equation}
Do the above differences really vanish? We answer this affirmatively and so proves McKay's theorem, based on the following simple lemma on group action.

\begin{lemma}\label{lem-action}
    Suppose $G$ is a finite group acting transitively on $\{1,2,\dots,n\}$. Then there is an orbit of pairs $\{(g\cdot i, g\cdot j),g\in G\}$ such that $\{\boldsymbol{e}_{g\cdot i}-\boldsymbol{e}_{g\cdot j}\}$ contains a basis of $\boldsymbol{1}^\perp=\{x\in\mathbb{R}^n, x\perp\boldsymbol{1}\}$, where $\boldsymbol{e}_i$ is the vector with the $i$-th entry $1$ and 0 elsewhere.
\end{lemma}

\begin{proof}
    It suffices to consider the orbit of $(1,2)$. Since $G$ acts transitively, for any $k\geq 3$, there exists $g_k\in G$ such that $g_k\cdot1=k$. Thus, the orbit of $(1,2)$ contains $(3,g_3\cdot2),\dots,(n,g_n\cdot2)$ and $(1,2)$ itself. The supports of these $(n-1)$ vectors cover all indices from $1$ to $n$,  whence $\boldsymbol{e}_{1}-\boldsymbol{e}_2, \boldsymbol{e}_{k}-\boldsymbol{e}_{g_k\cdot 2}, k\geq 3$, form a basis of $\boldsymbol{1}^\perp$.
\end{proof}

\begin{proposition}\label{prop-vanish}
    Provided the notation above and McKay's condition in Theorem \ref{thm-McKay}, there exists a permutation $\tilde{p}$ such that $\tilde{p}\boldsymbol{A}_2|_{U_2^\perp}\tilde{p}^T=\boldsymbol{A}_1|_{U_1^\perp}$. 
\end{proposition}
\begin{proof}
    Now that $p\boldsymbol{A}_2|_{U_2^\perp}p^T$ and $\boldsymbol{A}_1|_{U_1^\perp}$ are similar and $p\mu_j=u_j$ so that $U_1^\perp=pU_2^\perp$, we only need to consider the transformation of $\boldsymbol{A}_i$ on the basis $x_j$ ($px_j$), whose support is contained in that of $u_j$ and is perpendicular to $u_j$. If the support of $u_j$ is $o_j=\{j_1,\dots,j_m\}$, an orbit of $\mathrm{Aut}(T_1)$ which acts transitively on $o_j$, then by Lemma \ref{lem-action}, those $x_j$ can be chosen such that their supports consist in an orbit of $(j_1,j_2)$. Thus, we only need to consider the transformation of $\boldsymbol{A}_1$ on $\boldsymbol{e}_{j_1}-\boldsymbol{e}_{j_2}$, and so for $\boldsymbol{A}_2$. But since $p\boldsymbol{A}_2|_{U_2^\perp}p^T$ and $\boldsymbol{A}_1|_{U_1^\perp}$ are similar by McKay's condition, by reordering the indices in orbits $o_j$, we can see that the transformation of $\boldsymbol{A}_1$ and $\boldsymbol{A}_2$ are identical. These reorderings within orbits and the permutation $p$ among orbits form a refined permutation $\tilde{p}$ which conjugate $\boldsymbol{A}_1$ and $\boldsymbol{A}_2$, i.e., $\tilde{p}\boldsymbol{A}_2|_{U_2^\perp}\tilde{p}^T=\boldsymbol{A}_1|_{U_1^\perp}$.
\end{proof}

Finally, we can prove McKay's Theorem \ref{thm-McKay}.

\begin{proof}[Proof of Theorem \ref{thm-McKay}]
    By Proposition \ref{prop-vanish} and (\ref{eq-perp}), we see that 
    \[\tilde{p}\boldsymbol{A}_2\tilde{p}^T-\boldsymbol{A}_1=\tilde{p}\boldsymbol{A}_2|_{U_2^\perp}\tilde{p}^T-\boldsymbol{A}_1|_{U_1^\perp}=\boldsymbol{0},\]
    that is, by reordering columns and rows (synchronously), $\boldsymbol{A}_2$ becomes $\boldsymbol{A}_1$. Thus $T_1$ and $T_2$ are isomorphic.
\end{proof}

\section{The utility of McKay's theorem}
\label{app:tm-analysis}
\subsection{McKay's theorem as an isometry between matrix algebras in the Frobenius norm}

We recall McKay's efficient criterion of determining trees by the adjacency and degree matrix altogether as follows.

\begin{theorem}\label{thm-McKay's}
	Two trees $T_1$ and $T_2$ are isomorphic if and only if for any two-variable real polynomial $p$, $p(\boldsymbol{A}(T_1), \boldsymbol{D}(T_1))$ and $p(\boldsymbol{A}(T_2), \boldsymbol{D}(T_2))$ are cospectral.
\end{theorem}

Note again that here the two-variable polynomials may not be seen as in $\mathbb{R}[x,y]$ since the variables may not commute with each other. The spectrum of a matrix is actually determined by the traces of its powers as follows.

\begin{lemma}\label{lem-traces}
For any matrix $\boldsymbol{M}\in \boldsymbol{M}_n(\mathbb{R})$, its spectrum is determined by the traces $tr(\boldsymbol{M}^i)$ for $i=1,\dots,n$.
\end{lemma}
\begin{proof}
    Suppose the spectrum of $\boldsymbol{M}$ consists of $\lambda_1,\dots,\lambda_n\in\mathbb{C}$. Then its characteristic polynomial is $p(x)=(x-\lambda_1)\cdots(x-\lambda_n)=x^n+\sum_{i=1}^n(-1)^ie_ix^{n-i}$ with $e_i$ the elementary symmetric polynomials in $\lambda_j$'s. By Newton's identities, $e_i$ are determined by the power sums $\sum_{j=1}^n\lambda_j^i=tr(\boldsymbol{M}^i), i=1,\dots,n$, so is $p(x)$. Hence, the lemma follows.
\end{proof}

Lemma \ref{lem-traces} hints that McKay's theorem may only concerns aboutthe  comparison of the traces of polynomials or monomials in the adjacency and degree matrices. Note that a monomial in the two matrices is just a binary word in them.

\begin{corollary}\label{cor-traces}
    Two trees $T_1$ and $T_2$ of $n$ vertices are isomorphic if and only if for any binary word $w$, $tr(w(\boldsymbol{A}_1,\boldsymbol{D}_1)=tr(w(\boldsymbol{A}_2,\boldsymbol{D}_2))$, where $w(\boldsymbol{A}_i,\boldsymbol{D}_i)\in \boldsymbol{M}_n(\mathbb{R})$ is the realization of $w$ in the adjacency matrix $\boldsymbol{A}_i$ and degree matrix $\boldsymbol{D}_i$.
\end{corollary}
\begin{proof}
    By Lemma \ref{lem-traces}, the spectrum of any $p(\boldsymbol{A}_i,\boldsymbol{D}_i)$ is determined by $tr(p(\boldsymbol{A}_i,\boldsymbol{D}_i)^j)$, which may be written as the linear combination of traces of monomials or binary words in $\boldsymbol{A}_i$ and $\boldsymbol{D}_i$, since $tr(\cdot)$ is linear. Thus if for any binary word $w$, $tr(w(\boldsymbol{A}_1,\boldsymbol{D}_1))=tr(w(\boldsymbol{A}_2,\boldsymbol{D}_2))$, then the condition of Theorem \ref{thm-McKay's} is satisfied so that $T_1$ is isomorphic to $T_2$.
\end{proof}

Note that trace naturally defines a metric on matrices, i.e. the Frobenius norm, by $tr(\boldsymbol{A}^T\boldsymbol{A})=\|\boldsymbol{A}\|_F^2$. More specifically, for any two-variable polynomial $p,q$, 
\begin{equation}\label{eq-norm}
    d_F^2(p(\boldsymbol{A},\boldsymbol{D}),q(\boldsymbol{A},\boldsymbol{D})):=\|p(\boldsymbol{A},\boldsymbol{D})-q(\boldsymbol{A},\boldsymbol{D})\|_F^2=tr((p(\boldsymbol{A},\boldsymbol{D})-q(\boldsymbol{A},\boldsymbol{D}))^T(p(\boldsymbol{A},\boldsymbol{D})-q(\boldsymbol{A},\boldsymbol{D}))),
\end{equation}
where the transposes $p(\boldsymbol{A},\boldsymbol{D})^T, q(\boldsymbol{A},\boldsymbol{D})^T$ are still polynomials in $\boldsymbol{A}$ and $\boldsymbol{D}$ since they are symmetric. Thus, the trace characterization above may be further translated as an isometry between matrix algebras under the Frobenius norm as follows. Let $\mathcal{A}=\mathcal{A}(T)=\langle \boldsymbol{A},\boldsymbol{D}\rangle$ be the matrix algebra generated by the adjacency matrix $\boldsymbol{A}$ and the degree matrix $\boldsymbol{D}$ of a tree $T$, whose elements are polynomials in $\boldsymbol{A}$ and $\boldsymbol{D}$. For any two trees $T_1$ and $T_2$ with $n$ vertices, let $f:\mathcal{A}(T_1)\rightarrow \mathcal{A}(T_2)$ be the canonical homomorphism via $f(p(\boldsymbol{A}_1,\boldsymbol{D}_1))=p(\boldsymbol{A}_2,\boldsymbol{D}_2)$. Then

\begin{theorem}\label{thm-isometry}
     With the notations above, McKay's theorem (Theorem \ref{thm-McKay's}) is equivalent to the following statement: $T_1$ is isomorphic to $T_2$ if and only if the canonical homomorphism $f$ is an isometry under the Frobenius norm.
\end{theorem}

\begin{proof}
    By Corollary \ref{cor-traces}, we only need to verify that $f$ being an isometry is equivalent to the trace equality therein. Apparently, the latter implies the former by the expression (\ref{eq-norm}). On the other hand, for isometry $f$ and two-variable $p$, then from
    \[d_F^2(p(\boldsymbol{A},\boldsymbol{D}),0)=\|p(\boldsymbol{A},\boldsymbol{D})\|^2=tr(p(\boldsymbol{A},\boldsymbol{D})^Tp(\boldsymbol{A},\boldsymbol{D}))\]
    and 
    \begin{align}
        d_F^2(p(\boldsymbol{A},\boldsymbol{D}),\boldsymbol{I}_n)&=tr(p(\boldsymbol{A},\boldsymbol{D})^Tp(\boldsymbol{A},\boldsymbol{D})-p(\boldsymbol{A},\boldsymbol{D})^T-p(\boldsymbol{A},\boldsymbol{D})+\boldsymbol{I}_n)\\
        &=tr(p(\boldsymbol{A},\boldsymbol{D})^Tp(\boldsymbol{A},\boldsymbol{D}))-2tr(p(\boldsymbol{A},\boldsymbol{D}))+n,
    \end{align}

    we deduce $tr(p(\boldsymbol{A}_1,\boldsymbol{D}_1))=tr(p(\boldsymbol{A}_2,\boldsymbol{D}_2))$, i.e., $f$ being an isometry implies the condition of McKay's theorem.
\end{proof}

    Theorem \ref{thm-isometry} hints that, to canonically check whether two trees are isomorphic, we may search for a sequence of binary words in the adjacency and degree matrices, which span a dense subset of all words under the Frobenius norm. Since the algebra $\langle \boldsymbol{A},\boldsymbol{D}\rangle$ is always finite dimensional, it suffices to find a sequence of words as its basis. We investigate this idea in full details in the next section.

\subsection{Properties of potential canonical words and the confinement of distance measure}

    Suppose that $w_1,\dots,w_k$ are the words in search for spanning (together with their powers) a dense subset of $\mathcal{A}=\langle \boldsymbol{A},\boldsymbol{D}\rangle$ with respect to Theorem \ref{thm-isometry} for two trees of $n$ vertices. To check whether $w_i(\boldsymbol{A}_1,\boldsymbol{D}_1)$ and $w_i(\boldsymbol{A}_2,\boldsymbol{D}_2)$ are cospectral, by Lemma \ref{lem-traces}, it is equivalent to check $tr(w_i(\boldsymbol{A}_1,\boldsymbol{D}_1)^j)=tr(w_i(\boldsymbol{A}_2,\boldsymbol{D}_2)^j)$ for $j=1,\dots, n$. To be more specific, we define an inner product on the algebra $\mathcal{A}$ by 
    \begin{equation}\label{eq-inner product}
        (x,y)=tr(x^Ty)=tr(xy^T)=tr(yx^T), \forall x,y \in \mathcal{A},
    \end{equation}
    which is indeed the inner product of $\mathbb{R}^{n^2}$ with $\boldsymbol{M}_n(\mathbb{R})$ treated as $n^2$ dimensional vectors. In this regard, a canonical sequence of words in search should span the algebra $\mathcal{A}$. In other words, their linear combinations of the sequence should contain an orthogonal basis. Suppose $w_1,\dots,w_k$ are the words in search. Then we expect that for (almost) all $x\in\mathcal{A}$, if $tr(w_i^mx^T)=0, i=1,\dots,k, m\leq n$, or equivalently $tr(w_i^mx)=0$ since $x^T\in\mathcal{A}$ due to the symmetry of $\boldsymbol{A}$ and $\boldsymbol{D}$, then $x=\boldsymbol{0}$. Note that for $m>n, w_i^m$ can be expressed as a linear combination of $w_i^j, j\leq n$ by Cayley-Hamilton's theorem.

    However, the inner product as of (\ref{eq-inner product}) is not entirely determined by the spectrum of $w_i$'s, but rather their conjugacy classes. Simply, for any invertible $\boldsymbol{M}\in\mathcal{A}$, we have \[tr(\boldsymbol{M}^{-1}w_i^mx\boldsymbol{M})=tr(\boldsymbol{M}^{-1}w_i^m\boldsymbol{M}\boldsymbol{M}^{-1}x\boldsymbol{M})=tr(\boldsymbol{M}^{-1}w_i^m\boldsymbol{M}y), \forall y\in\mathcal{A},\]
    where the spectra of $w_i^m$ and the conjugates are not distinguishable. This indicates that, if we only measure the sequence $w_i$ through computing their inner products induced by trace, even if the results are identical for both trees, we may only conclude that the two trees are ``conjugate" up to some invertible weights from a potential variation matrix $\boldsymbol{M}$ as above. For example, if $\boldsymbol{M}$ is not a permutation, we may only conclude that $\boldsymbol{A}_1=\boldsymbol{M}^{-1}\boldsymbol{A}_2\boldsymbol{M}$, i.e., one tree is only similar with another (vertex) weighted tree (up to some permutation). Fortunately, this latter possibility is ruled out by the other part of the distance measure as of (\ref{eq:ours_distance}) in section 4.1, namely,
    \[d_c(\mathbf{S}_{\boldsymbol{x}}, \mathbf{S}_{\boldsymbol{y}})= \lambda  d_{g} (\mathbf{S}_{\boldsymbol{x}}, \mathbf{S}_{\boldsymbol{y}})+ (1-\lambda )d_{t} (\mathbf{S}_{\boldsymbol{x}}, \mathbf{S}_{\boldsymbol{y}}),
\]
where $d_g(\mathbf{S}_{\boldsymbol{x}}, \mathbf{S}_{\boldsymbol{y}})$ is the hyperbolic geodesic distance to measure the dissimilarity of the vertex sets $\mathbf{S}_{\boldsymbol{x}}$ and $ \mathbf{S}_{\boldsymbol{y}}$ from a global perspective. Thus, when $d_g$ is small, the topological measure $d_t$ is determined by the sequence $w_i$'s with respect the inner product $(\cdot,\cdot)$ as in (\ref{eq-inner product}).

In summary, with the confinement of the distance measure $d_g$, a canonical sequence of words in search for spectral distinction of trees are supposed to satisfy the following.

\begin{proposition}\label{prop-traces}
    For any two trees $T_1$ and $T_2$ of $n$ vertices, a sequence of canonical words $w_1,\dots,w_k$ in the adjacency and degree matrices that contain a basis with respect to the inner product as of (\ref{eq-inner product}), or equivalently span a dense subset of $\mathcal{A}$ with respect to Theorem \ref{thm-isometry} whence McKay's theorem \ref{thm-McKay's}, should satisfy, if $x\in\mathcal{A}$ and $tr(w_i^mx)=0, \forall i\leq k, m\leq n$, then $x=\boldsymbol{0}$.
\end{proposition}

\subsection{Thue-Morse sequence as a canonical identifier}

In this section, we try to argue in favor of suggesting a canonical sequence of binary words given by the Thue-Morse sequence, which could be seen as the sequence of least redundancy. This resides essentially in its overlap freeness as explained as follows.

First, suppose for some word $w_i$ in the sequence, there is $w_i=ab^t$ for some subwords $a, b$ of $w_i$ and integer $t>1$. Then 
\[tr(w_ix)=tr(ab^tx)=tr(b^txa),\]
the last equality of with is due to the basic fact that $tr(xy)=tr(yx)$. Since $a,b$ are also words in $\boldsymbol{A}$ and $\boldsymbol{D}$, $a,b\in\mathcal{A}$, so is $xa$. Thus $tr(w_ix)=0$ is implied by $tr(b^ty)=0$ so that we would better replace $w_i$ by $b$ to reduce redundancy as indicated by Proposition \ref{prop-traces}. This suggest that \textbf{we should use words with as least powers of subwords as possible}.

Second, words in two letters is inevitable to contain powers of subwords as length increases. Actually, any binary words of length 4 must contain powers, which are
\begin{align*}
    \boldsymbol{A}\boldsymbol{D}\boldsymbol{A}\boldsymbol{D}, D\boldsymbol{A}\boldsymbol{D}\boldsymbol{A}, \boldsymbol{A}\boldsymbol{D}^2\boldsymbol{A}, \boldsymbol{A}^2\boldsymbol{D}^2, \boldsymbol{D}\boldsymbol{A}^2\boldsymbol{D}, \boldsymbol{D}^2\boldsymbol{A}^2,  \boldsymbol{A}^3\boldsymbol{D}, \boldsymbol{A}^2\boldsymbol{D}\boldsymbol{A}, \\ \boldsymbol{A}\boldsymbol{D}\boldsymbol{A}^2, \boldsymbol{D}\boldsymbol{A}^3, \boldsymbol{D}^3\boldsymbol{A}, \boldsymbol{D}^2\boldsymbol{A}\boldsymbol{D}, \boldsymbol{D}\boldsymbol{A}\boldsymbol{D}^2, \boldsymbol{A}\boldsymbol{D}^3, \boldsymbol{A}^4, \boldsymbol{D}^4.
\end{align*}

Note that $\boldsymbol{A}\boldsymbol{D}\boldsymbol{A}\boldsymbol{D}=(\boldsymbol{A}\boldsymbol{D})^2, \boldsymbol{D}\boldsymbol{A}\boldsymbol{D}\boldsymbol{A}=(\boldsymbol{D}\boldsymbol{A})^2$. (There is an infinite sequence of square-free words in three letters, see Theorem 1.6.2 of ~\citep{allouche2003automatic}.) The binary words with least powers as redundancy may be construted from the Thue-Morse sequence by the following.

\begin{theorem}[Theorem 1.6.1 of ~\citep{allouche2003automatic}]\label{thm-TM}
    The Thue-Morse infinite word is overlap free.
\end{theorem}

Here, denoting the Thue-Morse sequence by $\boldsymbol{t}_0=0, \boldsymbol{t}_1=01, \boldsymbol{t}_2=0110, \boldsymbol{t}_3=01101001, \dots, \boldsymbol{t}_n=\boldsymbol{t}_{n-1}\boldsymbol{t}_{n-1}^*$, the Thue-Morse infinite word is $\boldsymbol{t}=\boldsymbol{t}_0\boldsymbol{t}_1\boldsymbol{t}_2\cdots$ by concatenation. A word $w$ is \textit{overlap free}, if it can not be written as $w=acxcxcb$ with $a,b,x$ subwords of $w$ and $c$ a single letter. For example, the word ``banana" is an overlap while ``falafel" is not. Theorem \ref{thm-TM} verifies our statement that the Thue-Morse sequence provides the binary words with least redundancy as follows.

\begin{corollary}
    The Thue-Morse words $T_i$ are cube-free.
\end{corollary}

Moreover, if we confine the canonical construction of binary words to automatic sequences, i.e. words generated by a self-operating symbolic  homomorphism on binary words, the overlap free choice of words are uniquely left with the Thue-Morse sequence. More specifically, let $\Sigma_2^*$ be the set of binary words, say on $\boldsymbol{A}$ and $\boldsymbol{D}$, and $h: \Sigma_2^*\rightarrow \Sigma_2^*$ be a homomorphism, i.e. $h(w_1w_2)=h(w_1)h(w_2), \forall w_1,w_2\in\Sigma_2^*$. For example, let $\mu(D)=\boldsymbol{D}\boldsymbol{A}, \mu(A)=\boldsymbol{A}\boldsymbol{D}$, then the Thue-Morse words are the iterations $\mu^i(\boldsymbol{D})$. $\mu$ is called the Thue-Morse morphism. We call $h$ overlap free, if $h(w)$ is overlap free for any overlap free word $w$. Then

\begin{theorem}\label{thm-T3}
   For any homomorphism $h$ on binary words in $\boldsymbol{A}$ and $\boldsymbol{D}$, if $h(\boldsymbol{D}\boldsymbol{A})$ is not empty and $h(\boldsymbol{t}_3)=h(\boldsymbol{D}\boldsymbol{A}A\boldsymbol{D}\boldsymbol{A}D\boldsymbol{D}\boldsymbol{A})$ is overlap free, then there exist integer $k\geq 0$ such that $h=\mu^k$ or $h=E\circ\mu^k$ with $E$ the complement homomorphism with $E(\boldsymbol{D})=\boldsymbol{A}, E(\boldsymbol{A})=\boldsymbol{D}$. 
\end{theorem}

The theorem above is a particular useful part of Corollary 1.7.8 of ~\citep{allouche2003automatic}, which basically means that any overlap-free homomorphism is aligned with the Thue-Morse morphism after the third operation. By a self-operating symbolic choice of words, we mean the words $h^i(D)$ or $E\circ h^i(D)$. Thus, among choosing the self-operating symbolic binary words of least redundancy (overlap free), it suffices to utilize the first 4 Thue-Morse words, $\boldsymbol{t}_0=\boldsymbol{D}, \boldsymbol{t}_1=\boldsymbol{D}\boldsymbol{A}, \boldsymbol{t}_2=\boldsymbol{D}\boldsymbol{A}\boldsymbol{A}\boldsymbol{D}, \boldsymbol{t}_3=\boldsymbol{D}\boldsymbol{A}\boldsymbol{A}\boldsymbol{D}\boldsymbol{A}\boldsymbol{D}\boldsymbol{D}\boldsymbol{A}$, which explains our choice of the topological measure $d_t$ in practice.

In summary, we showed that McKay's theorem (Theorem \ref{thm-McKay's}) of spectral distinction of trees is equivalent to the canonical homomorphism $f$ as of Theorem \ref{thm-isometry} is an isometry under the Frobenius norm between the matrix algebras generated by the adjacency and degree matrices of two trees. This indicates that to find a sequence of canonical binary words (independent of choices of trees) in the adjacency and degree matrices, it suffices to find a sequence spanning an orthogonal basis of the matrix algebra with respect to the inner product induced by trace, which is further confined to overlap free words in the sense of least redundancy, provided that the distance measure of vertices is small. Then among all automatic sequences overlap free words generated by self operating homomorphisms, the Thue-Morse sequence is essentially the ultimate choice since any other overlap free homomorphism is a power of the Thue-Morse homomorphism (or its complement). Moreover, the first 4 Thue-Morse words are enough to determine the overlap freeness of an automatic sequence by Theorem \ref{thm-T3}, which are then utilized to construct the topological measure.  

% \appendix

% \section{Technical Appendices and Supplementary Material}
% Technical appendices with additional results, figures, graphs and proofs may be submitted with the paper submission before the full submission deadline (see above), or as a separate PDF in the ZIP file below before the supplementary material deadline. There is no page limit for the technical appendices.

%%%%%%%%%%%%%%%%%%%%%%%%%%%%%%%%%%%%%%%%%%%%%%%%%%%%%%%%%%%%
% \input{checklist}

\end{document}